\pdfoutput=1

\documentclass[10pt,twocolumn,letterpaper]{article}

\usepackage[pagenumbers]{cvpr}

\usepackage{amsthm}
\usepackage{multirow}
\usepackage{placeins}
\usepackage{tabularx,array}
\usepackage{colortbl}
\usepackage{pifont}
\definecolor{rowgray}{gray}{0.92}

\newcommand{\IMMA}{\textsc{Imma}}
\newcommand{\CondNum}{\textsc{CondNum}}

\definecolor{cvprblue}{rgb}{0.21,0.49,0.74}
\usepackage[pagebackref,breaklinks,colorlinks,allcolors=cvprblue]{hyperref}

\title{Model Immunization Cannot Reliably Protect Models from Harmful Adaptation}

\author{
Tong Li\thanks{Equal contribution. \quad \textsuperscript{\textdagger}Corresponding author.} \quad
Chenru Wang\footnotemark[1] \quad
Chris Thomas \quad
Shengwei An\footnotemark[2]\\
Virginia Tech\\
{\tt\small
\{tongli01, crwang, christhomas, swan\}@vt.edu}
}

\begin{document}
\maketitle
\begingroup
\renewcommand{\thefootnote}{}
\footnotetext{Code is available on \href{https://github.com/litong2358/immunization-under-misalignment}{GitHub}.}
\endgroup
\begin{abstract}
Efficient fine-tuning techniques have substantially reduced the cost of adapting open-weight models, enabling malicious actors to readily fine-tune them toward harmful capabilities. In response, model immunization modifies a pretrained checkpoint to resist subsequent adaptation toward designated harmful capabilities while preserving its utility and adaptability for benign tasks. Furthermore, a few immunization methods have sought to keep their protection robust to some changes in the attacker's adaptation. However, fine-tuning offers attackers many more low-cost options, and we observe that existing methods still substantially overfit to the attacker configuration they are optimized against. Such overfitting stays hidden when immunization is evaluated under, or near, this configuration, so the reported protection may give a false sense of security. To rigorously evaluate model immunization, we conduct experiments, varying one component of this configuration at a time: the harmful data, the adaptation strategy, or the model initialization. Across four immunization methods on text-to-image generation and image classification, with additional results on language models, these variations substantially degrade the protection, and in many cases the immunized model relearns the harmful capability nearly as well as the un-immunized model. Existing immunization therefore cannot reliably protect models, leaving them vulnerable to harmful adaptation.\looseness=-1

\end{abstract}
    
\section{Introduction}
\label{sec:intro}

\begin{figure*}[t]
\centering
\includegraphics[width=\textwidth]{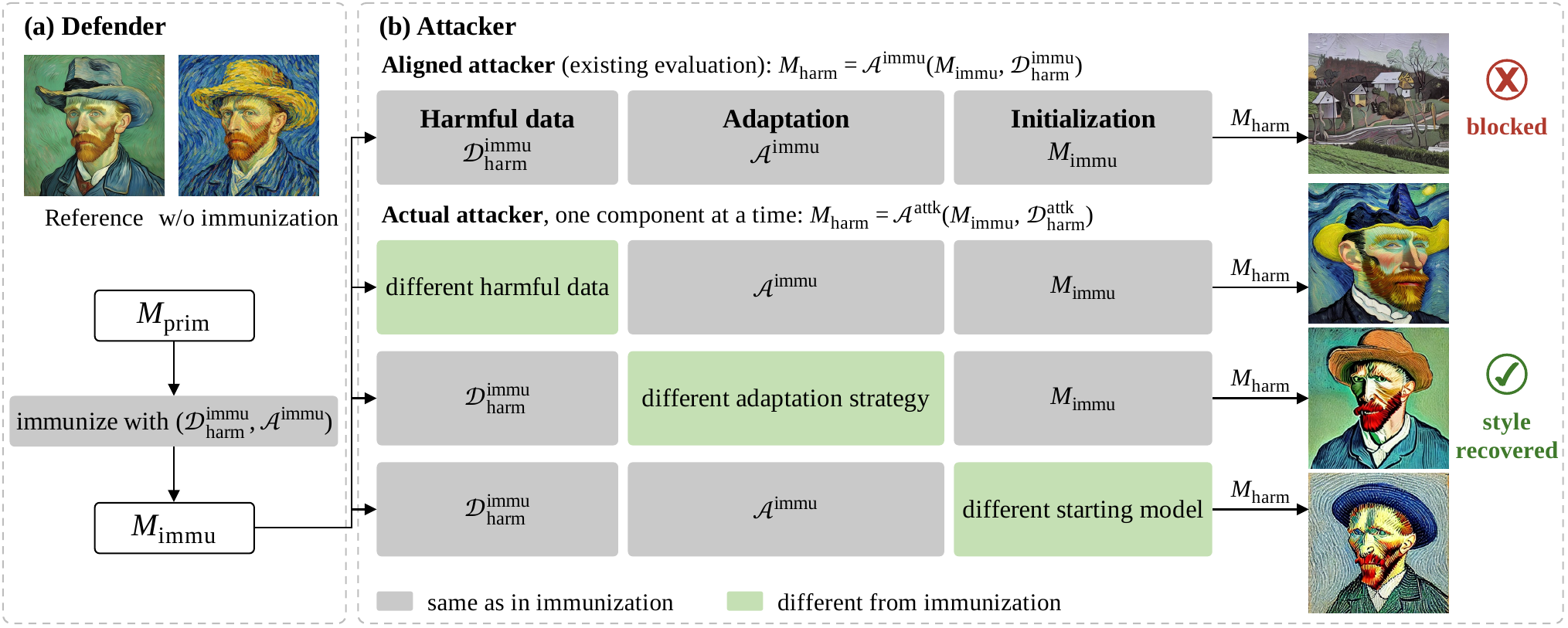}
\caption{Pipeline of model immunization and harmful adaptation. (a) The defender
immunizes $M_{\text{prim}}$ into $M_{\text{immu}}$ with
$\mathcal{D}^{\text{immu}}_{\text{harm}}$ and $\mathcal{A}^{\text{immu}}$. (b) Each row is an attacker,
specified by its harmful data, adaptation strategy, and initialization. Gray components
follow the configuration used during immunization, and green components depart from
it. The images are samples of the resulting $M_{\text{harm}}$, one per attacker.}
\label{fig:teaser}
\end{figure*}

Open-weight models are increasingly prevalent~\citep{kapoor2024}, and their value lies largely in their adaptability. With parameter-efficient fine-tuning~\citep{lora2022}, users can adapt a released checkpoint to downstream tasks at a fraction of the cost of training from scratch. However, the same flexibility also enables undesirable adaptations that model publishers may seek to prevent. For example, a text-to-image model can be fine-tuned to imitate an unauthorized artist's painting style~\citep{glaze2023,pham2024}, while a safety-aligned language model can be fine-tuned to exhibit behaviors it was originally trained to refuse~\citep{harmfulft2024,hftsurvey2024}. As illustrated in the top-left corner of Fig.~\ref{fig:teaser}, fine-tuning Stable Diffusion on a small collection of Vincent van Gogh's paintings enables the model to reproduce his distinctive artistic style (``w/o immunization''). Note that we use style imitation as an illustrative example throughout this paper, but the same concern applies to any capability that a model publisher seeks to withhold, and our evaluation also covers image classification, with additional results on language models.

To prevent such undesirable adaptations, recent studies have proposed model immunization~\citep{selfdestruct2023,imma2024,condnum2025,gift2025,clamp2026}, which modifies a model before its release to make subsequent harmful fine-tuning ineffective. These methods typically formulate immunization as a min-max optimization problem, in which the defender optimizes the model against a simulated attacker attempting to recover a targeted harmful capability. As shown by the aligned attacker in Fig.~\ref{fig:teaser}, immunization can successfully prevent the simulated fine-tuning attack from recovering Van Gogh's painting style.

However, the simulated attacker used during immunization need not represent the actual attacker encountered after model release. Once a checkpoint is publicly available, the defender can no longer constrain how it is modified. In particular, an actual attacker can use different fine-tuning data, alter the adaptation configuration, or modify the released checkpoint before fine-tuning. Encouragingly, \IMMA{}~\citep{imma2024} shows that its protection carries over between DreamBooth~\citep{dreambooth2023} and Textual Inversion~\citep{textualinversion2023}, and CLAMP~\citep{clamp2026} is designed to remain effective over thousands of fine-tuning steps. Nevertheless, fine-tuning offers an attacker many other low-cost ways to depart from the configuration used during immunization, beyond those examined so far. Although these deviations may appear minor, they can change the optimization trajectory and potentially circumvent protections learned against the simulated attacker. Indeed, the actual attacker in Fig.~\ref{fig:teaser} successfully recovers Van Gogh's painting style despite the model having been immunized. This observation raises a fundamental question: \emph{Does model immunization prevent harmful adaptation itself, or does it merely prevent the specific adaptation procedure anticipated during immunization?}

To answer this question, we systematically investigate the robustness of existing model immunization methods under attacker misalignment, where the actual attacker deviates from the attacker simulated during immunization. Specifically, we examine three attacker-controlled dimensions. First, for harmful data, we replace the defender's proxy dataset with independently collected samples and further apply model-agnostic or model-aware perturbations that preserve the targeted harmful capability while altering the fine-tuning trajectory. Second, for adaptation strategies, we vary fine-tuning configurations, such as learning rates, to explore optimization trajectories not anticipated during immunization. Third, for model initialization, we apply function-preserving reparameterizations to the released checkpoint, such as compensating rescaling of attention projections, to change its parameterization without altering its underlying function. Through these variations, we evaluate whether existing immunization methods provide robust protection against harmful adaptation or rely on close alignment between the simulated and actual attackers.\looseness=-1

We evaluate four representative immunization methods, \IMMA{}~\citep{imma2024}, MIMA~\citep{mima2025}, CLAMP~\citep{clamp2026}, and \CondNum{}~\citep{condnum2025}, on text-to-image generation and image classification. Our evaluation covers five artist styles and four classification benchmarks, and App.~\ref{app:llm} provides additional results for CLAMP on three language-model backbones. Our results reveal that existing immunization methods are substantially weakened across tasks when the actual attacker deviates from the simulated attacker used during immunization. In some cases, immunization loses its protective effect entirely: the immunized model relearns the targeted capability at least as quickly as its un-immunized counterpart. These findings suggest that existing immunization methods may provide a false sense of security, as their apparent effectiveness depends heavily on assumptions about the attacker's behavior.

Our contributions are summarized as follows:
\begin{itemize}

\item We identify and investigate a critical limitation of existing model immunization methods: their vulnerability to misalignment between the simulated attacker used during immunization and the actual attacker after model release.

\item We develop a systematic evaluation framework that examines attacker misalignment along three dimensions: harmful data, adaptation strategy, and model initialization.

\item We conduct extensive experiments across three task families and four immunization methods, demonstrating that even simple deviations from the simulated attacker can substantially weaken or completely defeat existing immunization protections.

\end{itemize}

\section{Related Work}
\label{sec:related}

\smallskip
\noindent
\textbf{Model Immunization.} \IMMA{}~\citep{imma2024} immunizes a text-to-image
diffusion model~\citep{ldm2022} against a simulated attacker adapting it via low-rank
adapters or DreamBooth~\citep{lora2022,dreambooth2023}, by moving the released weights
to raise that adaptation's loss. This bilevel displacement starts from a concept-erased
checkpoint~\citep{esd2023}. \CondNum{}~\citep{condnum2025} instead raises the condition
number of the harmful task's feature Gram matrix on a classification backbone~\citep{resnet2016}, slowing first-order optimization of a linear probe there while
keeping the pretraining task well-conditioned. CLAMP~\citep{clamp2026} instead reshapes
the optimization dynamics of gradient-based fine-tuning so that long-horizon adaptation
toward the harmful task stays contractive. This mechanism is not tied to one modality,
so the same procedure applies to a diffusion model, an image classifier, and a
safety-aligned language model. The
three sit in a wider release-time line that starts with task blocking, which meta-learns
weights whose adaptation to a harmful task fails~\citep{selfdestruct2023}. Further work
in this line immunizes several concepts at once~\citep{mima2025}, immunizes without
committing to the attacker's adaptation algorithm~\citep{gift2025}, obstructs few-shot
classification over restricted classes~\citep{lto2024}, makes a checkpoint
non-fine-tunable on restricted tasks~\citep{sophon2024}, locks an encoder against
probing on prohibited domains~\citep{encoderlock2025}, and states immunization
conditions for language models~\citep{immunizellm2024,hftsurvey2024}. \IMMA{},
\CondNum{}, and CLAMP each report an immunized-to-un-immunized ratio under their
evaluated attack configurations. We then evaluate these three methods, together with the multi-concept
MIMA~\citep{mima2025}, under a common protocol to measure what
the reported protection depends on.

\smallskip
\noindent
\textbf{Capability Removal and Protective Perturbations.} Immunization usually sits atop capability removal:
weight-level concept erasure~\citep{esd2023,uce2024}, inference-time steering~\citep{safelatent2023}, or classifier unlearning~\citep{salun2024}.
Fine-tuning is known
to reverse erasure alone~\citep{pham2024,imma2024}, which motivates immunization. We hold that checkpoint fixed
and vary only the immunization displacement. A parallel line instead has the
\emph{defender} perturb the \emph{images}, so models trained on them cannot acquire the
capability: unlearnable examples~\citep{unlearnable2021}, style-mimicry protection~\citep{glaze2023}, anti-personalization perturbations~\citep{antidreambooth2023}, and
edit-resistant images~\citep{photoguard2023}. Here the \emph{attacker} perturbs its own
images to promote learning, yet diagnostics transfer: availability attacks are known to
install a shortcut while leaving the underlying information intact~\citep{shortcuts2022}.
We interpret a perturbation that helps only on the immunized arm the same way.

\section{Threat Model}
\label{sec:threat}

\smallskip
\noindent
\textbf{Defender's Capabilities.}
The defender constructs the immunized model \(M_{\text{immu}}\) from an open-weight model $M_{\text{prim}}$ trained on the primary-task dataset \(\mathcal{D}_{\text{prim}}\), and therefore has white-box access to the model throughout the immunization process. The defender has access to \(\mathcal{D}_{\text{prim}}\) and controls the harmful proxy dataset \(\mathcal{D}_{\text{harm}}^{\text{immu}}\) and all configurations of the immunization procedure. The defender may also simulate known adaptation attacks $\mathcal{A}^{\text{immu}}$ during immunization. Once \(M_{\text{immu}}\) is released, however, the defender cannot control how downstream users access, modify, or adapt the model. In particular, the defender does not know the attacker’s adaptation dataset \(\mathcal{D}_{\text{harm}}^{\text{attk}}\) or adaptation strategy $\mathcal{A}^{\text{attk}}$, both of which may differ from those simulated during immunization.

\smallskip
\noindent
\textbf{Defender's Goals.}
An effective immunized model should satisfy the following three requirements:
1) \emph{Primary Utility:} The immunized model \(M_{\text{immu}}\) should preserve the original model's performance on the primary task, as measured on \(\mathcal{D}_{\text{prim}}\).
2) \emph{Harmful Inadaptability:} An attacker should be unable to derive a useful harmful model from \(M_{\text{immu}}\) using any feasible adaptation strategy. Formally, for an attacker-selected adaptation strategy \(\mathcal{A}^{\text{attk}}\) and harmful dataset \(\mathcal{D}_{\text{harm}}^{\text{attk}}\), the resulting model
$M_{\text{harm}} = \mathcal{A}^{\text{attk}}(M_{\text{immu}},\mathcal{D}_{\text{harm}}^{\text{attk}})$
should exhibit low performance on the designated harmful capability.
3) \emph{Benign Adaptability:} A benign user should remain able to adapt \(M_{\text{immu}}\) to tasks unrelated to the designated harmful capability. Specifically, given a benign adaptation strategy \(\mathcal{A}^{\text{beni}}\) and dataset \(\mathcal{D}_{\text{beni}}\), the resulting model
$M_{\text{beni}}= \mathcal{A}^{\text{beni}}(M_{\text{immu}},\mathcal{D}_{\text{beni}})$
should achieve high performance on the corresponding benign task,
ideally comparable to a model adapted from \(M_{\text{prim}}\).

\smallskip
\noindent
\textbf{Attacker's Capabilities.}
The attacker obtains the released immunized model \(M_{\text{immu}}\) and has white-box access to it. 
The attacker controls both the harmful adaptation dataset \(\mathcal{D}_{\text{harm}}^{\text{attk}}\) and the complete adaptation strategy \(\mathcal{A}^{\text{attk}}\), and may also perturb the released weights before adapting them. The attacker may also transform \(\mathcal{D}_{\text{harm}}^{\text{attk}}\) while preserving its harmful objective. Such transformations may be model-agnostic, such as additive random noise, or model-aware, such as gradient-guided perturbations.

\smallskip
\noindent
\textbf{Attacker's Goal.}
The attacker's goal is to obtain a model $M_{\text{harm}} = \mathcal{A}^{\text{attk}}(M_{\text{immu}},\mathcal{D}_{\text{harm}}^{\text{attk}})$ that has a similar performance as $\mathcal{A}^{\text{attk}}(M_{\text{prim}}, \mathcal{D}_{\text{harm}}^{\text{attk}})$ on the designated harmful capability.

\section{Attacker Misalignment in Immunization}
\label{sec:method}

As described in our threat model, the defender approximates the unknown post-release attacker during immunization using a harmful proxy dataset \(\mathcal{D}_{\text{harm}}^{\text{immu}}\) and a surrogate adaptation strategy \(\mathcal{A}^{\text{immu}}\). Existing evaluations examine selected variations in harmful data and adaptation strategies, but attackers have a broader range of low-cost options. Configuration-specific overfitting may therefore remain hidden beyond the tested settings. After \(M_{\text{immu}}\) is released, the defender cannot control either the data or the adaptation strategy selected by a downstream attacker. Consequently, there is no reason to assume that
$\mathcal{D}_{\text{harm}}^{\text{attk}}=\mathcal{D}_{\text{harm}}^{\text{immu}}$, $\mathcal{A}^{\text{attk}}=\mathcal{A}^{\text{immu}}$, or \(M_{\text{immu}}\) stays unperturbed before adaptation. We therefore examine this misalignment along each axis the attacker controls.

\subsection{Harmful-Data Misalignment}
\label{sec:method:a3}

We introduce three types of harmful-data misalignment: 1) cross-dataset misalignment, 2) model-agnostic perturbation, and 3) model-aware perturbation.

\smallskip
\noindent
\textbf{Cross-Dataset Misalignment.}
Ideally, the harmful capability targeted by immunization corresponds to an underlying concept or data distribution rather than to a particular finite dataset. 
However, because the defender cannot enumerate this distribution directly, they collect a finite proxy dataset
$\mathcal{D}_{\text{harm}}^{\text{immu}}=\{(x_i,y_i)\}_{i=1}^{n}\sim \mathcal{P}_{\text{harm}}$
and use it to simulate harmful adaptation during immunization. This dataset inevitably provides only a limited empirical representation of the harmful capability.

Existing immunizations are shown to be protective under the tested data settings. However, this evidence does not necessarily establish robustness to other data sources that represent the same harmful capability. In practice, a downstream attacker can independently collect another dataset
$\mathcal{D}_{\text{harm}}^{\text{attk}} =
\{(x'_j,y'_j)\}_{j=1}^{m}
\sim \mathcal{P}_{\text{harm}}$, but
$\mathcal{D}_{\text{harm}}^{\text{attk}}
\neq
\mathcal{D}_{\text{harm}}^{\text{immu}}$.
The two datasets target the same harmful capability but contain different samples and may exhibit different low-level characteristics.

To test this cross-dataset misalignment, we replace the defender's proxy dataset with an independently collected harmful dataset while keeping the adaptation strategy fixed. Specifically, we construct
$M_{\text{harm}}=\mathcal{A}^{\text{attk}}(M_{\text{immu}},\mathcal{D}_{\text{harm}}^{\text{attk}})$,
where \(\mathcal{D}_{\text{harm}}^{\text{attk}}\) represents the same harmful capability as \(\mathcal{D}_{\text{harm}}^{\text{immu}}\) but does not reuse its training samples.

This misalignment naturally applies across modalities. For image classifiers, the attacker can use different images representing the same harmful target classes. For language models, alternative instruction--response pairs can instantiate the same harmful capability.
For text-to-image models, alternative image--text pairs can represent the same prohibited concept.
Such multi-modal models further expose multiple sources of misalignment: the attacker can alter the visual component while preserving the textual component (image misalignment), alter the textual component while preserving the visual component (prompt misalignment), or change both components jointly. App.~\ref{app:a2} explains how we achieve textual misalignment for text-to-image models.

\smallskip
\noindent
\textbf{Model-Agnostic Perturbation on Images.}
Without accessing \(M_{\text{immu}}\), the attacker can apply standard image transformations. We consider two transformations: random noise \(\delta\sim\mathcal{U}(-\varepsilon,\varepsilon)\), followed by clipping \(x+\delta\) to the valid pixel range, and cropping followed by rescaling (App.~\ref{app:crop}). On classification, the random noise instead sets every pixel to $\pm\varepsilon$ with a random sign, matching the amplitude of the model-aware perturbation, as App.~\ref{app:method} details. These transformations preserve the harmful semantic content while altering low-level image statistics. Because they require no model weights, gradients, outputs, or knowledge of the immunization method, they represent a minimal-capability attack baseline.

\smallskip
\noindent
\textbf{Model-Aware Perturbation.}
Existing immunizations commonly modify $M_{\text{immu}}$ such that harmful fine-tuning on $\mathcal{D}_{\text{harm}}^{\text{immu}}$
produces misleading parameter gradients, such as pointing to the
opposite direction from successful harmful adaptation. A model-aware perturbation uses
this distorted gradient signal to perturb $\mathcal{D}_{\text{harm}}^{\text{attk}}$ 
so that fine-tuning $M_{\text{immu}}$ on the perturbed dataset induces gradients that are better aligned with a productive harmful-adaptation direction.
In continuous domains such as images, we apply PGA, while in discrete domains such as text, we search along the direction guided by the largest gradient. In both cases, the released weights stay fixed and only the attacker's input is optimized.
We next describe how we instantiate model-aware perturbation for diffusion models and image classifiers. Due to space constraints, we defer the corresponding instantiation for language models to App.~\ref{app:llm-method}.

\smallskip
\noindent
\ding{70} \textit{Perturbing images in diffusion models.}
For diffusion models, immunization modifies the local loss landscape around \(M_{\text{immu}}\) such that harmful fine-tuning on a sample \(x\sim\mathcal{D}_{\text{harm}}^{\text{attk}}\) follows an ineffective optimization trajectory within an immunization-induced basin. To test whether adaptation can leave this basin, we construct a bounded perturbation \(\delta\) that maximizes the diffusion model's noise-prediction loss:
\begin{equation}
  \delta^{*}=\arg\max_{\|\delta\|_{\infty}\leq\varepsilon}\mathcal{L}_{\text{diff}}(M_{\text{immu}};x+\delta),
  \label{eq:pga-diff}
\end{equation}
where \(\varepsilon\) bounds the perturbation magnitude. The resulting sample \(x+\delta^{*}\) preserves the harmful concept represented by \(x\) but induces a larger prediction error in \(M_{\text{immu}}\). Fine-tuning on the perturbed sample therefore produces a stronger and qualitatively different gradient signal, pulling the optimization trajectory away from the immunization-induced basin and enabling the model to relearn the harmful concept.

\smallskip
\noindent
\ding{70} \textit{Perturbing images in image classifiers.}
For image classifiers, the attacker trains a linear probe on the frozen features \(\Phi_{M_{\text{immu}}}(x)\), and immunization concentrates these features in a few directions so that the probe learns the harmful task slowly. We therefore consider a bounded perturbation \(\delta\) that maximizes both the accuracy of the attacker's probe and the spread of the features:
\begin{equation}
  \delta^{*}=\arg\max_{\|\delta\|_{\infty}\leq\varepsilon}\widehat{\mathrm{Acc}}\bigl(W(\delta)\bigr)+\beta\log\mathrm{PR}\bigl(\Phi_{M_{\text{immu}}}(x+\delta)\bigr),
  \label{eq:pga-cls}
\end{equation}
where \(W(\delta)\) is the probe obtained from a differentiable simulation of the attacker's training on the perturbed features (App.~\ref{app:cls-ggp}), \(\widehat{\mathrm{Acc}}\) is its accuracy on clean images, \(\mathrm{PR}\) is the participation ratio that measures the effective number of feature directions, and \(\beta\) balances the two terms. The first term steers \(\delta\) toward inputs from which the probe learns the harmful classes, and the second spreads the features back over many directions. Training the probe on the perturbed features can therefore recover the harmful task that immunization was meant to suppress.

\subsection{Adaptation and Initialization Misalignment}
\label{sec:method:adapt}

\smallskip
\noindent
\textbf{Adaptation Misalignment.}
For diffusion and language models, both immunization and the simulated attacker
\(\mathcal{A}^{\text{immu}}\) used during immunization typically rely on low-rank
adaptation (LoRA), a widely adopted parameter-efficient fine-tuning method.
However, the adaptation configuration assumed during immunization need not match
the one used by the actual attacker. We instantiate this misalignment by changing
only the learning rate of the attacker's fine-tuning, for example from
\(10^{-4}\) to \(10^{-3}\). The larger step moves the adapter farther from the
released weights, potentially leaving the region in which immunization
suppresses harmful adaptation. App.~\ref{app:rank} varies the LoRA rank instead.

\smallskip
\noindent
\textbf{Model Initialization Misalignment.}
The protection of immunization typically resides in the parameters of
\(M_{\text{immu}}\), and evaluation adapts these parameters exactly as released. In practice, however, the
attacker can modify the released parameters before adaptation, so adaptation
starts from a different initialization. We instantiate this misalignment by
rescaling the attention layers of the diffusion model without changing the
function it computes. An attention layer computes
\(Y=\mathrm{softmax}(QK^{\top}/\sqrt{d_k})\,VW_O\), where the query \(Q\), key
\(K\), and value \(V\) are projections of the layer input with weights \(W_Q\),
\(W_K\), and \(W_V\), \(d_k\) is the key dimension, and \(W_O\) is the output
projection. The attacker multiplies \(W_Q\) and \(W_O\) by a factor \(c\) and
divides \(W_K\) and \(W_V\) by \(c\), so \(QK^{\top}\) and \(VW_O\), and therefore
\(Y\), are unchanged. Since LoRA adds its update to the rescaled weights, the same
update acts \(c\) times more strongly through \(W_K\) and \(W_V\) and \(c\) times
more weakly through \(W_Q\) and \(W_O\). We thus evaluate whether the protection
persists for a model with the same function as \(M_{\text{immu}}\) but different
parameters.

\section{Experiments}
\label{sec:exp}

\subsection{Setup}
\label{sec:exp:setup}

\smallskip
\noindent
\textbf{Immunizations and Models.}
We evaluate four immunizations using their original configurations.
\IMMA{}~\citep{imma2024} and MIMA~\citep{mima2025} are evaluated on Stable Diffusion v1.4~\citep{ldm2022},
\CondNum{}~\citep{condnum2025} on a ResNet-18 classifier~\citep{resnet2016}, and
CLAMP~\citep{clamp2026} on both architectures. App.~\ref{app:llm} further evaluates CLAMP on three language models. For each immunization, we use either the official implementation or the checkpoints provided by the authors.

\smallskip
\noindent
\textbf{Datasets.} The diffusion experiments cover
five artist styles: Van~Gogh, Tyler~Edlin, Kilian~Eng, Thomas~Kinkade, and
Kelly~McKernan. The classification experiments cover four harmful
classification tasks: 1) Stanford Cars~\citep{cars2013}, a $196$-category
fine-grained car-model recognition task, 2) Food-101~\citep{food2014}, a
$101$-category dish recognition task, 3) Flowers-102~\citep{flowers2008}, a
$102$-category flower recognition task, and 4) VMMRdb-196~\citep{vmmrdb2017}, a $196$-category vehicle make-and-model recognition task
drawn from photos collected from the internet.

\smallskip
\noindent
\textbf{Implementation.}
Unless otherwise specified, we follow the adaptation strategies used in the original immunization studies, which we refer to as \emph{plain adaptation}. For PGA on images, we constrain the perturbation to \(\lVert\delta\rVert_{\infty}\leq 1/255\) against \CondNum{} and \(\lVert\delta\rVert_{\infty}\leq 2/255\) against CLAMP for image classification, and \(\lVert\delta\rVert_{\infty}\leq 4/255\) for diffusion models. We use the same perturbation budgets for the corresponding model-agnostic controls.
Additional details are provided in App.~\ref{app:method} and ~\ref{app:instantiation}.

\subsection{Evaluation Metrics}
\label{sec:exp:metrics}

We use the same metrics defined in the literature to evaluate the immunization effects across different model families. Similarity Gap Ratio (SGR) was proposed by \IMMA{}~\citep{imma2024} to measure the immunization gain on diffusion models and later extended to classification and language models~\citep{clamp2026}. The general definition of SGR is:
\begin{equation}
  \mathrm{SGR}(M_{\text{prim}}, M_{\text{immu}})=\frac{U(M_{\text{prim}})-U(M_{\text{immu}})}{U(M_{\text{prim}})} \times 100,
  \label{eq:sgr}
\end{equation}
where $U(M)$ measures the adaptability of a model $M$, that is, the harmful-task performance of a model adapted from $M$. Intuitively, an effective immunized model should have a lower adaptability and thus a high SGR. In contrast, the attacker desires a lower SGR. A lower SGR
means that less protection remains, while a negative value means that the
immunized model achieves higher harmful-task adaptability than the original model.

For \textit{diffusion models}, we measure adaptability by the similarity between images generated by the adapted model and the reference. Following \IMMA{}, we average the $\mathrm{SGR}$ of the similarities computed by CLIP~\citep{clip2021}, DINO~\citep{dino2021}, and
$1-\mathrm{LPIPS}$~\citep{lpips2018}.
For \textit{classification models}, we measure adaptability by classification accuracy. We further report $\mathrm{ACC}_O$ and $\mathrm{ACC}_I$ that denote the accuracy of the model adapted from
$M_{\text{prim}}$ and $M_{\text{immu}}$, respectively.

\subsection{Results on Diffusion Models}
\label{sec:exp:diffusion}

\begin{table*}[t]
\centering
\caption{Main generation results. Average $\mathrm{SGR}$; lower is
better for the attacker. Combined applies image misalignment,
prompt misalignment, and model-aware PGA together. The model-agnostic row is the matched
control for model-aware PGA. Bold marks the lowest $\mathrm{SGR}$ for each defense and artist.}
\label{tab:gen-main}
\footnotesize
\renewcommand{\arraystretch}{1.15}
\setlength{\tabcolsep}{2pt}
\definecolor{rowgray}{gray}{0.92}
\newcommand{\ctl}[1]{\textcolor{black!55}{#1}}
\begin{tabularx}{\textwidth}{ll*{5}{>{\centering\arraybackslash}X}}
\toprule
Defense & Attack & Van Gogh & T. Edlin & K. Eng & T. Kinkade & K. McKernan \\
\midrule
\IMMA{} & Plain adaptation & $17.63_{\pm 2.55}$ & $18.62_{\pm 2.26}$ & $21.80_{\pm 5.62}$ & $18.43_{\pm 1.19}$ & $23.57_{\pm 0.36}$ \\
\specialrule{0.9pt}{1.5pt}{1.5pt}
& Image misalignment & $15.52_{\pm 2.28}$ & $17.13_{\pm 2.98}$ & $10.28_{\pm 5.19}$ & $11.35_{\pm 2.65}$ & $12.15_{\pm 2.72}$ \\
& Prompt misalignment & $\mathbf{6.46}_{\pm 1.20}$ & $\mathbf{8.98}_{\pm 1.40}$ & $11.84_{\pm 1.99}$ & $9.13_{\pm 2.40}$ & $9.26_{\pm 2.78}$ \\
& Model-aware (PGA) & $8.41_{\pm 0.33}$ & $13.93_{\pm 0.37}$ & $11.15_{\pm 2.92}$ & $13.07_{\pm 2.59}$ & $16.27_{\pm 0.42}$ \\
& \ctl{Model-agnostic (noise)} & \ctl{$14.08_{\pm 0.41}$} & \ctl{$14.82_{\pm 1.75}$} & \ctl{$13.50_{\pm 2.37}$} & \ctl{$17.01_{\pm 2.35}$} & \ctl{$21.95_{\pm 3.63}$} \\
\rowcolor{rowgray}
& Combined & $\mathbf{6.46}_{\pm 2.07}$ & $12.91_{\pm 1.65}$ & $\mathbf{6.80}_{\pm 1.44}$ & $\mathbf{6.23}_{\pm 3.16}$ & $\mathbf{-1.53}_{\pm 1.57}$ \\
\midrule
CLAMP & Plain adaptation & $18.41_{\pm 1.09}$ & $40.08_{\pm 0.06}$ & $21.47_{\pm 3.24}$ & $30.52_{\pm 4.54}$ & $37.18_{\pm 0.69}$ \\
\specialrule{0.9pt}{1.5pt}{1.5pt}
& Image misalignment & $12.78_{\pm 3.35}$ & $32.40_{\pm 1.15}$ & $11.34_{\pm 0.80}$ & $19.96_{\pm 2.15}$ & $\mathbf{29.88}_{\pm 7.55}$ \\
& Prompt misalignment & $7.88_{\pm 2.64}$ & $36.45_{\pm 0.63}$ & $16.56_{\pm 3.69}$ & $24.32_{\pm 8.58}$ & $32.91_{\pm 3.43}$ \\
& Model-aware (PGA) & $10.05_{\pm 0.60}$ & $35.06_{\pm 3.96}$ & $17.38_{\pm 1.34}$ & $21.74_{\pm 2.41}$ & $33.44_{\pm 3.16}$ \\
& \ctl{Model-agnostic (noise)} & \ctl{$14.56_{\pm 2.40}$} & \ctl{$38.49_{\pm 0.77}$} & \ctl{$20.24_{\pm 3.92}$} & \ctl{$24.33_{\pm 2.59}$} & \ctl{$34.01_{\pm 2.77}$} \\
\rowcolor{rowgray}
& Combined & $\mathbf{7.07}_{\pm 0.87}$ & $\mathbf{17.93}_{\pm 2.30}$ & $\mathbf{6.40}_{\pm 4.30}$ & $\mathbf{14.23}_{\pm 3.81}$ & $31.92_{\pm 2.87}$ \\
\bottomrule
\end{tabularx}
\end{table*}
\begin{figure*}[t]
\centering
\includegraphics[width=\textwidth]{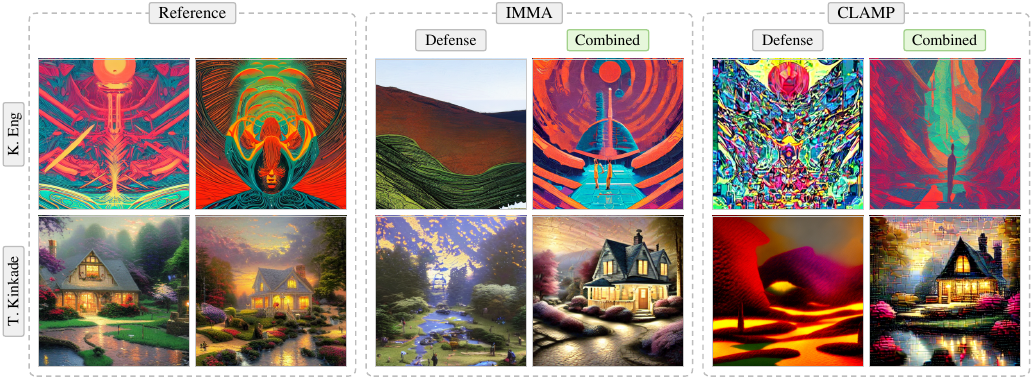}
\caption{Adapted samples for Kilian~Eng and Thomas~Kinkade. \emph{Defense} adapts
$M_{\text{immu}}$ with plain adaptation, and \emph{Combined} with the combined harmful-data misalignments of Tab.~\ref{tab:gen-main}.
The \emph{Defense} samples miss the \emph{Reference} style, and the \emph{Combined} samples recover it.}
\label{fig:qualitative}
\end{figure*}

\begin{figure*}[t]
\centering
\includegraphics{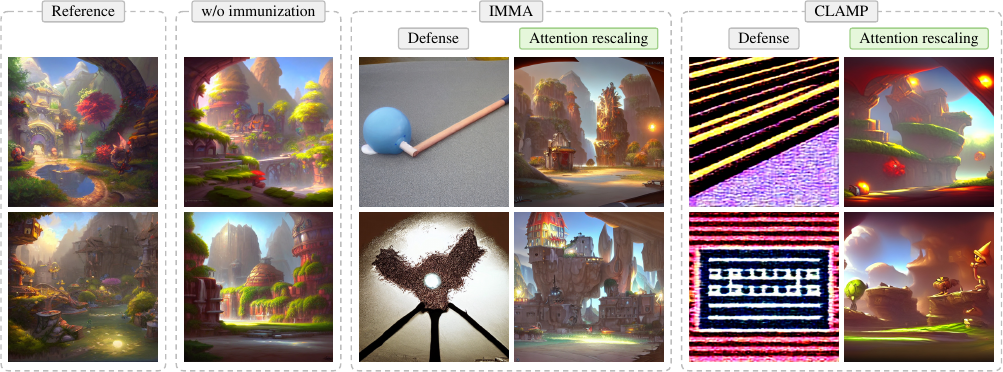}
\caption{Adapted samples for Tyler~Edlin under model initialization misalignment, with attention
weights rescaled by $c=10$. The \emph{Defense} samples lose the artist's style, and the
\emph{Attention rescaling} samples bring it back.}
\label{fig:init-samples}
\end{figure*}

\smallskip
\noindent
\textbf{Harmful-Data Misalignment.} Tab.~\ref{tab:gen-main} reports
each harmful-data misalignment and their
combination, where $\pm$ denotes the standard deviation.
Cross-dataset misalignment, which replaces the images (Image misalignment)
or rewrites only the prompt (Prompt misalignment), model-agnostic noise, and
model-aware PGA each lower $\mathrm{SGR}$ relative to plain adaptation in
every defense--artist cell. An attacker may also apply several harmful-data
misalignments simultaneously. The Combined row does so and lowers $\mathrm{SGR}$ in every
cell. For example, $\mathrm{SGR}$ decreases from $23.57$ to $-1.53$ on \IMMA{} with Kelly~McKernan. By definition of $\mathrm{SGR}$, this negative value means the immunized model is even more vulnerable than the un-immunized one in this case. 
Fig.~\ref{fig:qualitative} shows qualitative samples, in which the protected style reappears under misalignment. More results are provided in App.~\ref{app:gen-full}.

\smallskip
\noindent
\textbf{Adaptation Misalignment.} We show the protection of both defenses
when the attacker changes the learning rate $\eta$ of its fine-tuning. As
Tab.~\ref{tab:lr} shows, both \IMMA{} and CLAMP struggle once the attacker's
learning rate departs from the $10^{-4}$ used in their original evaluations,
and $\mathrm{SGR}$ decreases steadily as $\eta$ grows for all five artists.
Moreover, this reduction comes from $M_{\text{immu}}$ reaching a higher
similarity. Beyond the learning rate, we also change the attacker's LoRA rank and
observe the same trend.

\begin{table}[!t]
\centering
\caption{Learning-rate misalignment on diffusion (SGR). $\eta$ is the attacker's
learning rate, and plain adaptation uses $\eta=10^{-4}$. $U(M_{\text{prim}})$ is measured
at the same $\eta$.}
\label{tab:lr}
\footnotesize
\renewcommand{\arraystretch}{1.15}
\setlength{\tabcolsep}{2pt}
\begin{tabular}{llccc}
\toprule
Defense & Artist & Plain adapt. & $\eta{=}3{\times}10^{-4}$ & $\eta{=}10^{-3}$ \\
\midrule
\IMMA{} & Van Gogh    & $17.63_{\pm 2.55}$ & $6.46_{\pm 5.24}$ & $\mathbf{5.09}_{\pm 3.88}$ \\
        & T. Edlin    & $18.62_{\pm 2.26}$ & $4.73_{\pm 3.48}$ & $\mathbf{4.02}_{\pm 3.04}$ \\
        & K. Eng      & $21.80_{\pm 5.62}$ & $6.97_{\pm 3.75}$ & $\mathbf{3.89}_{\pm 2.55}$ \\
        & T. Kinkade  & $18.43_{\pm 1.19}$ & $5.36_{\pm 4.99}$ & $\mathbf{0.75}_{\pm 2.98}$ \\
        & K. McKernan & $23.57_{\pm 0.36}$ & $10.49_{\pm 2.65}$ & $\mathbf{3.54}_{\pm 2.88}$ \\
\midrule
CLAMP   & Van Gogh    & $18.41_{\pm 1.09}$ & $8.72_{\pm 5.05}$ & $\mathbf{5.44}_{\pm 1.31}$ \\
        & T. Edlin    & $40.08_{\pm 0.06}$ & $33.92_{\pm 7.88}$ & $\mathbf{6.70}_{\pm 11.89}$ \\
        & K. Eng      & $21.47_{\pm 3.24}$ & $12.45_{\pm 5.38}$ & $\mathbf{1.17}_{\pm 1.74}$ \\
        & T. Kinkade  & $30.52_{\pm 4.54}$ & $19.38_{\pm 9.91}$ & $\mathbf{8.76}_{\pm 6.86}$ \\
        & K. McKernan & $37.18_{\pm 0.69}$ & $28.60_{\pm 7.48}$ & $\mathbf{23.76}_{\pm 3.68}$ \\
\bottomrule
\end{tabular}
\vspace{-15pt}
\end{table}

\begin{table}[t]
\centering
\caption{Model-initialization misalignment on diffusion (SGR). $c$ is the factor by
which the attacker rescales the attention weights before adaptation, and plain
adaptation corresponds to $c=1$. For $c=3$ and $c=10$, $U(M_{\text{prim}})$ is
measured after rescaling $M_{\text{prim}}$ by the same $c$. Bold marks the lowest
value in each row.}
\label{tab:init}
\footnotesize
\renewcommand{\arraystretch}{1.15}
\setlength{\tabcolsep}{2.5pt}
\begin{tabular}{llccc}
\toprule
Defense & Artist & Plain adapt. & $c=3$ & $c=10$ \\
\midrule
\IMMA{} & Van Gogh    & $17.63_{\pm 2.55}$ & $13.33_{\pm 4.23}$ & $\mathbf{7.36}_{\pm 5.69}$ \\
        & T. Edlin    & $18.62_{\pm 2.26}$ & $14.44_{\pm 2.99}$ & $\mathbf{7.65}_{\pm 3.68}$ \\
        & K. Eng      & $21.80_{\pm 5.62}$ & $18.96_{\pm 4.30}$ & $\mathbf{7.87}_{\pm 4.19}$ \\
        & T. Kinkade  & $18.43_{\pm 1.19}$ & $14.08_{\pm 3.48}$ & $\mathbf{4.81}_{\pm 4.50}$ \\
        & K. McKernan & $23.57_{\pm 0.36}$ & $18.86_{\pm 4.42}$ & $\mathbf{12.91}_{\pm 0.51}$ \\
\midrule
CLAMP   & Van Gogh    & $18.41_{\pm 1.09}$ & $12.19_{\pm 2.63}$ & $\mathbf{5.26}_{\pm 4.29}$ \\
        & T. Edlin    & $40.08_{\pm 0.06}$ & $39.02_{\pm 4.99}$ & $\mathbf{16.85}_{\pm 7.53}$ \\
        & K. Eng      & $21.47_{\pm 3.24}$ & $12.59_{\pm 3.95}$ & $\mathbf{1.96}_{\pm 3.08}$ \\
        & T. Kinkade  & $30.52_{\pm 4.54}$ & $19.48_{\pm 1.37}$ & $\mathbf{7.01}_{\pm 1.61}$ \\
        & K. McKernan & $37.18_{\pm 0.69}$ & $34.82_{\pm 2.82}$ & $\mathbf{16.91}_{\pm 3.29}$ \\
\bottomrule
\end{tabular}
\end{table}

\smallskip
\noindent
\textbf{Model Initialization Misalignment.} Before adaptation, the attacker
rescales the attention weights of $M_{\text{immu}}$ by a factor $c$ in a way that
leaves the released model's outputs unchanged.
Tab.~\ref{tab:init} shows that both $c=3$ and $c=10$ lower $\mathrm{SGR}$ on both
defenses and all five artists, and that $c=10$ gives the lowest value in all ten
defense--artist cells. At $c=10$ it drops from $18.43$ to $4.81$ on \IMMA{} and
from $30.52$ to $7.01$ on CLAMP with Thomas~Kinkade.
Fig.~\ref{fig:init-samples} shows qualitative samples for Tyler~Edlin at $c=10$.

\smallskip
\noindent
\textbf{Multi-Concept Immunization.} We further evaluate MIMA, which immunizes
several concepts at once, on four groups of two or three artists. We apply the same prompt misalignment and learning-rate change
as above. Tab.~\ref{tab:mima} shows that both misalignments lower $\mathrm{SGR}$ in
all four groups. Moreover, at $\eta=10^{-3}$, $\mathrm{SGR}$ falls below zero in every
group, where adapting $M_{\text{immu}}$ reaches a higher similarity than adapting
$M_{\text{prim}}$ for every artist of the group. App.~\ref{app:mima} gives the setup
and further results.

\begin{table}[t]
\centering
\caption{Multi-concept immunization with MIMA ($\mathrm{SGR}$, averaged over the artists of each group).
$\eta$ is the attacker's learning rate. Groups 1 to 4 are (K.~Eng, T.~Kinkade, K.~McKernan), (K.~McKernan, Van~Gogh), (K.~Eng, C.~Monet, T.~Edlin), and (K.~Eng, K.~McKernan).}
\label{tab:mima}
\footnotesize
\renewcommand{\arraystretch}{1.15}
\setlength{\tabcolsep}{6pt}
\begin{tabular}{lccc}
\toprule
 & Plain adapt. & Prompt & $\eta=10^{-3}$ \\
\midrule
Group 1 & $25.60_{\pm 8.24}$ & $12.17_{\pm 2.21}$ & $\mathbf{-5.07}_{\pm 4.11}$ \\
Group 2 & $16.61_{\pm 6.62}$ & $9.36_{\pm 2.38}$ & $\mathbf{-7.06}_{\pm 4.76}$ \\
Group 3 & $15.61_{\pm 3.22}$ & $5.17_{\pm 1.94}$ & $\mathbf{-6.82}_{\pm 2.94}$ \\
Group 4 & $25.27_{\pm 5.73}$ & $6.09_{\pm 1.84}$ & $\mathbf{-2.25}_{\pm 3.88}$ \\
\midrule
Mean & $20.77$ & $8.20$ & $\mathbf{-5.30}$ \\
\bottomrule
\end{tabular}
\end{table}

\subsection{Results on Classification Models}
\label{sec:exp:classification}

\begin{table*}[t]
\centering
\caption{Classification results against \CondNum{} and CLAMP. $\mathrm{ACC}_I$ and
$\mathrm{ACC}_O$ (\%) are averaged over epochs $10/20/30/40/50$, and
$\Delta\mathrm{ACC}_I$ is the relative change of $\mathrm{ACC}_I$ from plain adaptation.
Perturbations are bounded by $\lVert\delta\rVert_\infty\leq1/255$ against \CondNum{} and
$2/255$ against CLAMP. The $\mathrm{SGR}$ of each attack uses as un-immunized reference
$M_{\text{prim}}$ adapted on the same perturbed images. App.~\ref{app:attack-o} reports
$\mathrm{ACC}_O$ under the two attacks.}
\label{tab:a3-cn}
\footnotesize
\setlength{\tabcolsep}{2.5pt}
\begin{tabular}{ll ccc ccc ccc}
\toprule
& & \multicolumn{3}{c}{Plain adaptation} & \multicolumn{3}{c}{Model-agnostic (noise)} & \multicolumn{3}{c}{Model-aware (PGA)} \\
\cmidrule(lr){3-5}\cmidrule(lr){6-8}\cmidrule(lr){9-11}
Defense & Dataset
& $\mathrm{SGR}$ & $\mathrm{ACC}_I$ & $\mathrm{ACC}_O$
& $\mathrm{SGR}$ & $\mathrm{ACC}_I$ & $\Delta\mathrm{ACC}_I$
& $\mathrm{SGR}$ & $\mathrm{ACC}_I$ & $\Delta\mathrm{ACC}_I$ \\
\midrule
\CondNum{} & Stanford Cars
& $20.99_{\pm 5.90}$ & $22.67_{\pm 1.38}$ & $28.28_{\pm 0.32}$
& $9.39_{\pm 3.38}$ & $25.70_{\pm 0.69}$ & $+13.38\%$
& $\mathbf{-2.25}_{\pm 1.68}$ & $28.16_{\pm 0.35}$ & $+24.22\%$ \\
& Food-101
& $4.65_{\pm 1.68}$ & $45.52_{\pm 0.75}$ & $47.67_{\pm 0.25}$
& $4.83_{\pm 2.17}$ & $45.46_{\pm 0.96}$ & $-0.15\%$
& $\mathbf{2.53}_{\pm 0.80}$ & $46.54_{\pm 0.28}$ & $+2.22\%$ \\
& Flowers-102
& $5.61_{\pm 0.99}$ & $71.15_{\pm 0.53}$ & $75.16_{\pm 0.33}$
& $5.87_{\pm 0.54}$ & $71.18_{\pm 0.74}$ & $+0.04\%$
& $\mathbf{-1.43}_{\pm 1.15}$ & $75.88_{\pm 0.85}$ & $+6.65\%$ \\
& VMMRdb-196
& $35.58_{\pm 3.01}$ & $7.78_{\pm 0.50}$ & $11.72_{\pm 0.39}$
& $33.80_{\pm 4.29}$ & $7.94_{\pm 0.56}$ & $+2.16\%$
& $\mathbf{26.55}_{\pm 5.59}$ & $9.17_{\pm 0.69}$ & $+17.94\%$ \\
\midrule
CLAMP & Stanford Cars
& $38.05_{\pm 1.65}$ & $9.90_{\pm 0.21}$ & $15.91_{\pm 0.22}$
& $38.61_{\pm 1.56}$ & $9.52_{\pm 0.20}$ & $-3.77\%$
& $\mathbf{16.95}_{\pm 0.88}$ & $13.32_{\pm 0.13}$ & $+34.58\%$ \\
& Food-101
& $5.15_{\pm 0.25}$ & $27.99_{\pm 0.17}$ & $29.48_{\pm 0.18}$
& $5.08_{\pm 0.48}$ & $28.22_{\pm 0.17}$ & $+0.83\%$
& $\mathbf{2.95}_{\pm 0.52}$ & $33.05_{\pm 0.16}$ & $+18.10\%$ \\
& Flowers-102
& $20.58_{\pm 0.53}$ & $33.54_{\pm 0.30}$ & $42.71_{\pm 0.61}$
& $22.30_{\pm 0.40}$ & $32.15_{\pm 0.18}$ & $-4.13\%$
& $\mathbf{4.78}_{\pm 1.90}$ & $39.32_{\pm 0.36}$ & $+17.24\%$ \\
& VMMRdb-196
& $13.23_{\pm 0.35}$ & $4.24_{\pm 0.27}$ & $5.27_{\pm 0.11}$
& $10.33_{\pm 2.69}$ & $4.50_{\pm 0.16}$ & $+6.06\%$
& $\mathbf{-8.80}_{\pm 3.19}$ & $6.50_{\pm 0.25}$ & $+53.39\%$ \\
\bottomrule
\end{tabular}
\vspace*{-5mm}
\end{table*}

\begin{figure}[t]
\centering
\includegraphics[width=\columnwidth]{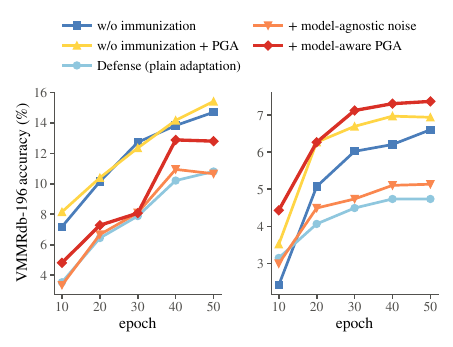}
\caption{Probe accuracy (\%) per epoch on VMMRdb-196 against \CondNum{} (left)
and CLAMP (right). The two \emph{w/o immunization} lines adapt
$M_{\text{prim}}$ on the clean and on the PGA-perturbed images, and the other
three lines adapt $M_{\text{immu}}$. Averaging each line over the five epochs
gives its accuracy in Tab.~\ref{tab:a3-cn} or Tab.~\ref{tab:attk-acco}.}
\label{fig:cls-curves}
\end{figure}

We apply PGA against the frozen feature extractor of each defense and train the same
linear probe on the perturbed features. Model-agnostic noise at the same
budget serves as the control. Tab.~\ref{tab:a3-cn} reports that, against both \CondNum{} and CLAMP, model-aware PGA raises $\mathrm{ACC}_I$ above
both plain adaptation and the model-agnostic control on every dataset. On some datasets, $\mathrm{ACC}_I$ with PGA is higher than $\mathrm{ACC}_O$.
Fig.~\ref{fig:cls-curves} shows that on VMMRdb-196 this gain over plain adaptation
holds at every recorded epoch.
App.~\ref{app:a1} further shows that immunizing \CondNum{} on synthetic images also
lowers $\mathrm{SGR}$.
Beyond the harmful data, we also change the learning rate and the optimizer of the attacker's
adaptation on classification, and App.~\ref{app:lr-cls} shows that both lower $\mathrm{SGR}$.

\subsection{Ablation Study and Adaptive Defense}
\label{sec:exp:ablation}

\smallskip
\noindent
\textbf{Model-Aware versus Model-Agnostic Perturbation.} Each model-aware
perturbation is compared with a matched control at the same budget, isolating
how much of the effect requires the gradients of $M_{\text{immu}}$. On
diffusion models, model-aware PGA reaches a lower $\mathrm{SGR}$ than
model-agnostic noise in all ten defense--artist cells. On classification, it does so on all four datasets under both defenses and
drives $\mathrm{SGR}$ below zero on Stanford Cars and Flowers-102 under \CondNum{} and on VMMRdb-196 under CLAMP, where the
control remains positive.

\smallskip
\noindent
\textbf{Adaptive Defender.} 
In practice, a defender may anticipate attacker misalignment and adapt the immunization procedure accordingly.
\begin{table}[t]
\centering
\vspace*{9.84pt}
\caption{Adaptive defender on diffusion ($\mathrm{SGR}$). \IMMA{} is immunized with five prompts that
contain the artist's name.}
\label{tab:adaptive-defender}
\footnotesize
\renewcommand{\arraystretch}{1.15}
\setlength{\tabcolsep}{6pt}
\begin{tabular}{lcc}
\toprule
Artist & Plain adaptation & Prompt misalignment \\
\midrule
Van Gogh    & $20.04_{\pm 5.50}$ & $6.48_{\pm 1.56}$ \\
T. Edlin    & $17.17_{\pm 1.83}$ & $10.00_{\pm 2.72}$ \\
K. Eng      & $19.37_{\pm 5.87}$ & $12.80_{\pm 1.85}$ \\
T. Kinkade  & $15.23_{\pm 1.22}$ & $4.41_{\pm 0.14}$ \\
K. McKernan & $22.08_{\pm 1.98}$ & $10.31_{\pm 1.89}$ \\
\bottomrule
\end{tabular}
\end{table}
To counter prompt misalignment, the defender may immunize the model using multiple prompts rather than a single one. In Tab.~\ref{tab:adaptive-defender}, we evaluate this strategy by jointly using five prompts containing the artist's name, such as \texttt{A painting by <Artist>}, during immunization. Plain adaptation, which uses one of these five prompts, maintains its original $\mathrm{SGR}$, indicating that immunization remains effective against the aligned attacker. However, prompt misalignment still reduces $\mathrm{SGR}$ across all five artists. These results demonstrate that even when the defender anticipates prompt variations and incorporates multiple prompts during immunization, the resulting protection remains vulnerable to prompt misalignment. Likewise, to counter learning-rate misalignment, the defender may immunize with the attacker's learning rate. App.~\ref{app:adaptive-lr} shows that matching the attacker's learning rate does not restore the protection either. Note that developing more advanced immunization is beyond the scope of this work and will be our future work.

\section{Conclusion}
\label{sec:conclusion}

In this work, we show that existing immunization methods remain vulnerable to
configuration-specific overfitting when the actual attacker departs from the
settings considered during immunization. We then evaluate existing immunization under such misalignment by
varying the harmful data, the adaptation strategy, and the model initialization
separately. Across four immunization methods and three modalities, these variations
substantially weaken the protection. In certain cases, they even remove it entirely,
and the immunized model relearns the harmful capability as well as the un-immunized
one. Moreover, immunization that is tailored to the anticipated prompt or
learning-rate misalignment remains vulnerable to it. Existing immunization therefore cannot
reliably protect released models from harmful adaptation. These findings highlight the
need for immunization that remains robust beyond the attacker configuration
anticipated by the defender.

{
    \small
    \bibliographystyle{ieeenat_fullname}
    \bibliography{main}
}

\clearpage
\maketitlesupplementary

\appendix
\setcounter{dbltopnumber}{5}
\raggedbottom

\section*{Appendix Contents}
\begin{itemize}\setlength{\itemsep}{2pt}
  \renewcommand{\texorpdfstring}[2]{#2}%
  \item Appendix~\ref{app:method}: \nameref{app:method} \dotfill \pageref{app:method}
  \item Appendix~\ref{sec:threat:criteria}: \nameref{sec:threat:criteria} \dotfill \pageref{sec:threat:criteria}
  \item Appendix~\ref{app:instantiation}: \nameref{app:instantiation} \dotfill \pageref{app:instantiation}
  \item Appendix~\ref{app:a1}: \nameref{app:a1} \dotfill \pageref{app:a1}
  \item Appendix~\ref{app:a2}: \nameref{app:a2} \dotfill \pageref{app:a2}
  \item Appendix~\ref{app:a3}: \nameref{app:a3} \dotfill \pageref{app:a3}
  \item Appendix~\ref{app:llm}: \nameref{app:llm} \dotfill \pageref{app:llm}
  \item Appendix~\ref{sec:disc}: \nameref{sec:disc} \dotfill \pageref{sec:disc}
\end{itemize}

\section{Attack construction details}
\label{app:method}

This appendix provides the implementation details omitted from
Sec.~\ref{sec:method} and Sec.~\ref{sec:exp}.

\paragraph{Image misalignment.}
The attacker replaces the twenty SD-v1.4-generated images with twenty images
sampled from SDXL~\citep{sdxl2024} for the same prompt. The immunized model, artist name
prompt, LoRA configuration, and optimization schedule remain fixed, so the
generator that produced the fine-tuning images is the only variable. The
substitute set is used for fine-tuning only; similarity is scored against the
unchanged SD-v1.4 reference set. App.~\ref{app:a1} gives the generation
protocol.

\paragraph{Prompt misalignment.}
The image files and immunized model remain unchanged. We replace
\texttt{An artwork by <Artist>} with a name-free style description.
Tab.~\ref{tab:a2-prompts} lists the exact descriptions.

\paragraph{Crop-and-resize.}
\label{app:crop}
This model-agnostic perturbation takes a randomly positioned
$384\times384$ square crop of each $512\times512$ training image, keeping
$75\%$ of each side, and the trainer rescales it back to $512\times512$
bilinearly. Nothing
else changes, and the released model is never queried. On Kilian~Eng it lowers
$\mathrm{SGR}$ from $21.80$ to $14.65$ under \IMMA{} and from $21.47$ to
$15.62$ under CLAMP (Tab.~\ref{tab:crop}).

\begin{table}[ht]
\centering
\caption{Crop-and-resize on Kilian~Eng. Mean $\mathrm{SGR}$;
lower is better for the attacker.}
\label{tab:crop}
\vspace{4pt}
\small
\setlength{\tabcolsep}{4pt}
\begin{tabular}{llc}
\toprule
Defense & Attack & $\mathrm{SGR}$ \\
\midrule
\IMMA{} & Plain adaptation & $21.80_{\pm 5.62}$ \\
        & Crop $75\%$      & $\mathbf{14.65}_{\pm 7.23}$ \\
\midrule
CLAMP   & Plain adaptation & $21.47_{\pm 3.24}$ \\
        & Crop $75\%$      & $\mathbf{15.62}_{\pm 5.48}$ \\
\bottomrule
\end{tabular}
\end{table}

\paragraph{Model initialization misalignment (attention rescaling).}
\label{app:attn-rescale}
Before adaptation, the attacker rescales all $32$ attention layers of the released
U-Net, the self- and cross-attention layers of its $16$ transformer blocks: the query
and output projections are multiplied by $c$, and the key and value projections are
divided by $c$. The rescaled model produces the same outputs as the released one, and
the same rescaling is applied to $M_{\text{prim}}$ when measuring $U_O$. The LoRA
adaptation, its schedule, and the evaluation are unchanged. Tab.~\ref{tab:init-du}
reports the change in style similarity relative to $c=1$, which is plain adaptation.
$\Delta U_O$ stays near zero or falls, between $-0.036$ and $-0.001$, while
$\Delta U_I$ rises in every cell, between $+0.002$ and $+0.133$. The rescaling
therefore mainly lifts the immunized model.

\begin{table*}[t]
\centering
\caption{Change in style similarity under attention rescaling, relative to plain
adaptation ($c=1$) and averaged over the three metrics. $\Delta U_O$ and $\Delta U_I$
are the changes in the similarity reached after adapting $M_{\text{prim}}$ and
$M_{\text{immu}}$.}
\label{tab:init-du}
\small
\renewcommand{\arraystretch}{1.15}
\setlength{\tabcolsep}{6pt}
\begin{tabular}{lllcccccc}
\toprule
Defense & $c$ & & Van Gogh & T. Edlin & K. Eng & T. Kinkade & K. McKernan & Mean \\
\midrule
\IMMA{}  & $3$  & $\Delta U_O$ & $-0.002$ & $-0.013$ & $-0.001$ & $-0.005$ & $-0.015$ & $-0.007$ \\
         &      & $\Delta U_I$ & $+0.023$ & $+0.018$ & $+0.039$ & $+0.026$ & $+0.021$ & $+0.026$ \\
         & $10$ & $\Delta U_O$ & $-0.015$ & $-0.036$ & $-0.017$ & $-0.014$ & $-0.021$ & $-0.021$ \\
         &      & $\Delta U_I$ & $+0.016$ & $+0.044$ & $+0.071$ & $+0.079$ & $+0.056$ & $+0.053$ \\
\midrule
CLAMP    & $3$  & $\Delta U_O$ & $-0.002$ & $-0.013$ & $-0.001$ & $-0.005$ & $-0.015$ & $-0.007$ \\
         &      & $\Delta U_I$ & $+0.035$ & $+0.002$ & $+0.053$ & $+0.073$ & $+0.007$ & $+0.034$ \\
         & $10$ & $\Delta U_O$ & $-0.015$ & $-0.036$ & $-0.017$ & $-0.014$ & $-0.021$ & $-0.021$ \\
         &      & $\Delta U_I$ & $+0.062$ & $+0.133$ & $+0.064$ & $+0.114$ & $+0.069$ & $+0.088$ \\
\bottomrule
\end{tabular}
\end{table*}

\begin{figure*}[t]
\centering
\includegraphics[width=\textwidth]{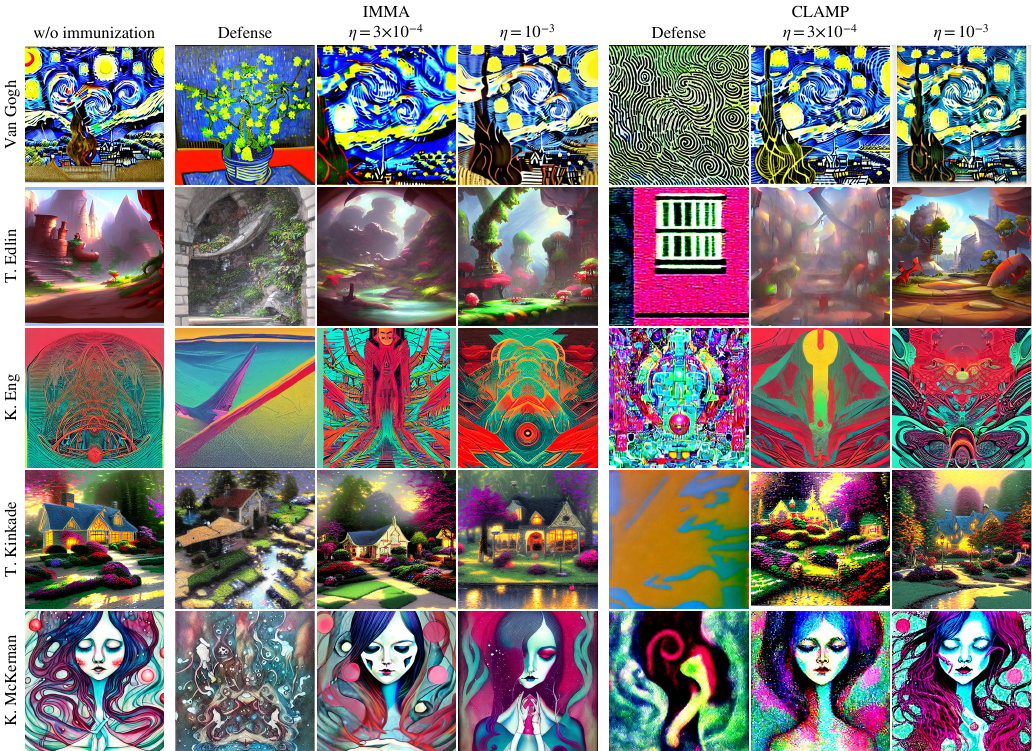}
\caption{Adapted samples under adaptation misalignment with learning rate $\eta$. The
\emph{Defense} samples use $\eta=10^{-4}$ and lose the artist's style, and the samples
at higher $\eta$ bring it back.}
\label{fig:lr-samples}
\end{figure*}

\begin{figure*}[t]
\centering
\includegraphics[width=\textwidth]{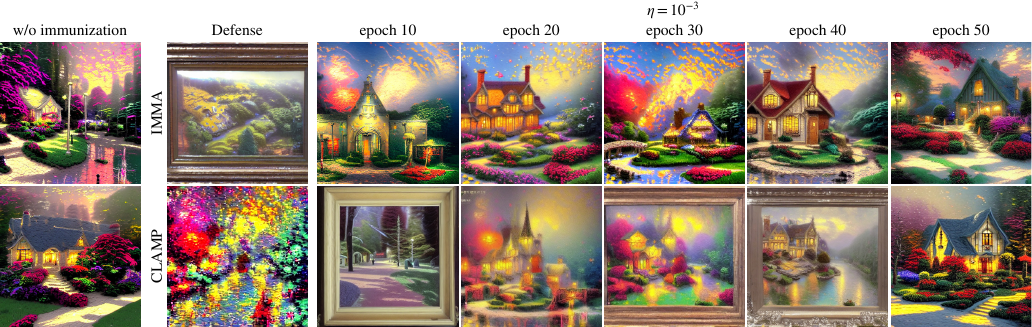}
\caption{Adapted samples of Thomas~Kinkade over fine-tuning under learning-rate misalignment.
\emph{w/o immunization} and \emph{Defense} use plain adaptation ($\eta=10^{-4}$), and the
remaining columns adapt $M_{\text{immu}}$ with $\eta=10^{-3}$. The \emph{Defense} samples
lose the artist's style, while with $\eta=10^{-3}$ the style reappears as fine-tuning
proceeds.}
\label{fig:lr-progress}
\end{figure*}

\paragraph{Adaptation misalignment (learning rate).}
\label{app:lr}
The attacker changes only the constant learning rate $\eta$ of its LoRA fine-tuning,
from $10^{-4}$ to $3\times10^{-4}$ or $10^{-3}$. The rank-$4$ adapter, the optimizer,
the number of epochs, the images, the prompt, and the evaluation are unchanged, and
$M_{\text{prim}}$ is adapted with the same $\eta$ when measuring $U_O$.
Tab.~\ref{tab:lr-du} reports the change in style similarity relative to
$\eta=10^{-4}$, which is plain adaptation. $\Delta U_O$ stays between $-0.024$ and
$+0.002$, while $\Delta U_I$ rises in every cell, between $+0.038$ and $+0.209$. The
higher learning rate therefore mainly lifts the immunized model, while the
un-immunized reference barely moves. Fig.~\ref{fig:lr-samples} shows adapted samples
for the five artists. Fig.~\ref{fig:lr-progress} follows one of them over
fine-tuning.

\begin{table*}[t]
\centering
\caption{Change in style similarity under learning-rate misalignment, relative to plain
adaptation ($\eta=10^{-4}$) and averaged over the three metrics. $\Delta U_O$ and $\Delta U_I$
are the changes in the similarity reached after adapting $M_{\text{prim}}$ and
$M_{\text{immu}}$.}
\label{tab:lr-du}
\small
\renewcommand{\arraystretch}{1.15}
\setlength{\tabcolsep}{6pt}
\begin{tabular}{lllcccccc}
\toprule
Defense & $\eta$ & & Van Gogh & T. Edlin & K. Eng & T. Kinkade & K. McKernan & Mean \\
\midrule
\IMMA{}  & $3\times10^{-4}$  & $\Delta U_O$ & $-0.001$ & $-0.008$ & $+0.001$ & $-0.001$ & $-0.009$ & $-0.003$ \\
         &                   & $\Delta U_I$ & $+0.062$ & $+0.088$ & $+0.094$ & $+0.088$ & $+0.082$ & $+0.083$ \\
         & $10^{-3}$         & $\Delta U_O$ & $+0.001$ & $-0.024$ & $-0.004$ & $+0.002$ & $-0.021$ & $-0.009$ \\
         &                   & $\Delta U_I$ & $+0.072$ & $+0.079$ & $+0.108$ & $+0.123$ & $+0.117$ & $+0.100$ \\
\midrule
CLAMP    & $3\times10^{-4}$  & $\Delta U_O$ & $-0.001$ & $-0.008$ & $+0.001$ & $-0.001$ & $-0.009$ & $-0.003$ \\
         &                   & $\Delta U_I$ & $+0.055$ & $+0.038$ & $+0.056$ & $+0.077$ & $+0.054$ & $+0.056$ \\
         & $10^{-3}$         & $\Delta U_O$ & $+0.001$ & $-0.024$ & $-0.004$ & $+0.002$ & $-0.021$ & $-0.009$ \\
         &                   & $\Delta U_I$ & $+0.077$ & $+0.209$ & $+0.118$ & $+0.151$ & $+0.079$ & $+0.127$ \\
\bottomrule
\end{tabular}
\end{table*}

\paragraph{Adaptive defender with the learning rate.}
\label{app:adaptive-lr}
A defender may also anticipate the learning-rate misalignment. We set the learning rate of
$\mathcal{A}^{\text{immu}}$ to the attacker's $\eta=3\times10^{-4}$ and rerun both
immunizations, after which the attacker adapts the resulting models at the same $\eta$.
As Tab.~\ref{tab:adaptive-lr} shows, $\mathrm{SGR}$ stays below that of plain adaptation
for both defenses and all five artists, and falls to less than half of it in all but one
case. Matching the attacker's learning rate therefore does not restore the protection.

\begin{table}[t]
\centering
\caption{Adaptive defender with the learning rate ($\mathrm{SGR}$). The attacker fine-tunes with
$\eta=3{\times}10^{-4}$, and the immunization uses the same $\eta$.}
\label{tab:adaptive-lr}
\footnotesize
\renewcommand{\arraystretch}{1.15}
\setlength{\tabcolsep}{6pt}
\begin{tabular}{lcc}
\toprule
Artist & \IMMA{} & CLAMP \\
\midrule
Van Gogh    & $3.65_{\pm 1.43}$ & $4.60_{\pm 3.29}$ \\
T. Edlin    & $5.68_{\pm 2.52}$ & $32.15_{\pm 7.32}$ \\
K. Eng      & $8.38_{\pm 3.97}$ & $8.91_{\pm 3.51}$ \\
T. Kinkade  & $4.06_{\pm 2.23}$ & $11.70_{\pm 2.43}$ \\
K. McKernan & $8.22_{\pm 2.49}$ & $16.28_{\pm 8.91}$ \\
\bottomrule
\end{tabular}
\end{table}

\paragraph{Learning-rate misalignment on classification.}
\label{app:lr-cls}
The CLAMP attacker adapts the immunized model with a base learning rate of
$\eta=10^{-2}$. We raise this base $\eta$ by the same factors of $3$
and $10$ as on diffusion. Tab.~\ref{tab:lr-cls} shows that $\mathrm{SGR}$ falls
from $38.05$ to $23.19$ on Stanford Cars and from $20.58$ to $6.87$ on Flowers-102.
For \CondNum{}, whose analysis assumes an optimal step size, the probe in our
evaluation already uses the best learning rate for each dataset.

\begin{table}[t]
\centering
\caption{Learning-rate misalignment on CLAMP classification (SGR). $\eta$ is the
base learning rate of the attacker's adaptation, and $\mathrm{ACC}_O$ is measured at the
same $\eta$.}
\label{tab:lr-cls}
\footnotesize
\renewcommand{\arraystretch}{1.15}
\setlength{\tabcolsep}{4pt}
\begin{tabular}{lccc}
\toprule
Dataset & Plain adapt. & $\eta{=}3{\times}10^{-2}$ & $\eta{=}10^{-1}$ \\
\midrule
Stanford Cars & $38.05_{\pm 1.65}$ & $28.50_{\pm 1.81}$ & $\mathbf{23.19}_{\pm 0.61}$ \\
Flowers-102   & $20.58_{\pm 0.53}$ & $12.85_{\pm 0.47}$ & $\mathbf{6.87}_{\pm 0.38}$ \\
\bottomrule
\end{tabular}
\end{table}

\paragraph{Optimizer misalignment on classification.}
\label{app:optim-cls}
We evaluate \CondNum{} against an attacker that trains its linear probe with
Adam~\cite{kingma2015adam} instead of SGD, so that each parameter has its own
step size. Tab.~\ref{tab:optim-cls} shows that $\mathrm{SGR}$ falls on all four
datasets, for example from $5.61$ to $0.93$ on Flowers-102 and from $35.58$ to
$10.09$ on VMMRdb-196. The protection of \CondNum{} thus weakens once the attacker
adapts with a different optimizer.

\begin{table}[t]
\centering
\caption{Optimizer misalignment on \CondNum{} classification. Adam replaces the
SGD of plain adaptation, and $\mathrm{ACC}_O$ is measured with the same
optimizer. Bold marks the lower $\mathrm{SGR}$ for each dataset.}
\label{tab:optim-cls}
\footnotesize
\renewcommand{\arraystretch}{1.15}
\setlength{\tabcolsep}{3pt}
\begin{tabular}{llccc}
\toprule
Dataset & Adaptation & $\mathrm{SGR}$ & $\mathrm{ACC}_I$ & $\mathrm{ACC}_O$ \\
\midrule
Stanford Cars & Plain & $20.99_{\pm 5.90}$ & $22.67_{\pm 1.38}$ & $28.28_{\pm 0.32}$ \\
              & Adam  & $\mathbf{14.54}_{\pm 0.80}$ & $30.44_{\pm 0.22}$ & $35.60_{\pm 0.09}$ \\
\midrule
Food-101      & Plain & $4.65_{\pm 1.68}$ & $45.52_{\pm 0.75}$ & $47.67_{\pm 0.25}$ \\
              & Adam  & $\mathbf{2.92}_{\pm 0.26}$ & $48.09_{\pm 0.05}$ & $49.55_{\pm 0.07}$ \\
\midrule
Flowers-102   & Plain & $5.61_{\pm 0.99}$ & $71.15_{\pm 0.53}$ & $75.16_{\pm 0.33}$ \\
              & Adam  & $\mathbf{0.93}_{\pm 0.26}$ & $77.84_{\pm 0.32}$ & $78.55_{\pm 0.12}$ \\
\midrule
VMMRdb-196    & Plain & $35.58_{\pm 3.01}$ & $7.78_{\pm 0.50}$ & $11.72_{\pm 0.39}$ \\
              & Adam  & $\mathbf{10.09}_{\pm 1.54}$ & $15.56_{\pm 0.18}$ & $17.26_{\pm 0.40}$ \\
\bottomrule
\end{tabular}
\end{table}

\begin{figure*}[t]
\centering
\includegraphics[width=\textwidth]{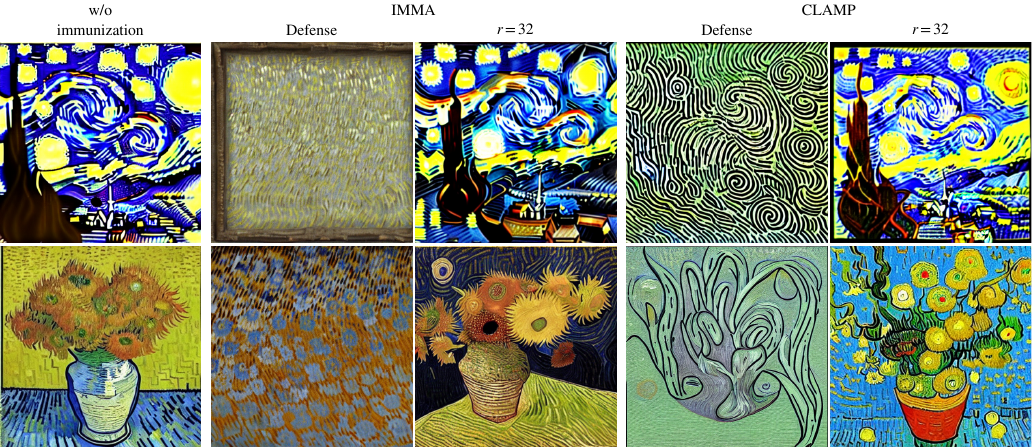}
\caption{Adapted samples for Van~Gogh under adaptation misalignment with LoRA
rank $r=32$. The \emph{Defense} samples lose the artist's style, and the
$r=32$ samples bring it back.}
\label{fig:rank-samples}
\end{figure*}

\paragraph{Adaptation misalignment (LoRA rank).}
\label{app:rank}
The attacker instead changes the LoRA rank $r$ of $\mathcal{A}^{\text{attk}}$ from $4$
to $8$, $16$, and $32$ and keeps its learning rate at $10^{-4}$. A higher-rank adapter
spans a larger update space, potentially exposing optimization directions that were
insufficiently constrained during immunization. $M_{\text{prim}}$ is adapted with the
same rank when measuring $U_O$. Tab.~\ref{tab:rank} shows that $\mathrm{SGR}$ falls as
$r$ grows on both defenses and both artists. At $r=32$ it drops from $23.57$ to $13.17$
on \IMMA{} with Kelly~McKernan and from $37.18$ to $29.69$ on CLAMP with
Kelly~McKernan. Fig.~\ref{fig:rank-samples} shows samples for Van~Gogh at $r=32$.

\begin{table}[t]
\centering
\caption{LoRA-rank misalignment on diffusion (SGR);
plain adaptation uses $r{=}4$.}
\label{tab:rank}
\footnotesize
\renewcommand{\arraystretch}{1.15}
\setlength{\tabcolsep}{3pt}
\definecolor{rowgray}{gray}{0.92}
\begin{tabular}{llcc}
\toprule
Defense & Attack & Van Gogh & K. McKernan \\
\midrule
\IMMA{} & Plain adapt. & $17.63_{\pm 2.55}$ & $23.57_{\pm 0.36}$ \\
     & $r=8$        & $14.76_{\pm 4.11}$ & $17.99_{\pm 3.92}$ \\
     & $r=16$       & $10.92_{\pm 3.78}$ & $17.40_{\pm 2.95}$ \\
     & $r=32$       & $\mathbf{8.72}_{\pm 2.26}$ & $\mathbf{13.17}_{\pm 3.37}$ \\
\midrule
CLAMP & Plain adapt. & $18.41_{\pm 1.09}$ & $37.18_{\pm 0.69}$ \\
      & $r=8$        & $18.20_{\pm 5.16}$ & $36.75_{\pm 1.46}$ \\
      & $r=16$       & $14.59_{\pm 2.24}$ & $34.57_{\pm 2.46}$ \\
      & $r=32$       & $\mathbf{11.01}_{\pm 2.86}$ & $\mathbf{29.69}_{\pm 1.73}$ \\
\bottomrule
\end{tabular}
\end{table}

\paragraph{Model-aware perturbation for classification.}
The attacker freezes the feature extractor of the immunized classifier and maximizes
the objective of Eq.~\ref{eq:a3-cn-obj} along the balanced direction $g$ of
Eq.~\ref{eq:a3-cf-vote}. Against \CondNum{}, the search runs for $40$ steps
subject to $\lVert\delta\rVert_\infty\leq1/255$. Against CLAMP,
the search iterates the projected step
$\delta \leftarrow \Pi_{\mathcal{C}_{\mathrm{pix}}}\bigl(\delta+\alpha\,\mathrm{sign}(g)\bigr)$
for $40$ steps with $\varepsilon=2/255$ and $\alpha=\varepsilon/4$. $\mathrm{PR}$ is computed on batches of $128$ images. Against CLAMP, both terms are
further averaged over four views of each image, redrawn at every step by random
cropping, horizontal flipping, and color jitter. Against \CondNum{}, each image enters
as is, without augmentation. Before feature
extraction, every perturbed image is rounded through uint8, so only changes
that survive ordinary image storage are credited. The model-agnostic control sets every
pixel to $\pm\varepsilon$ with a random sign. Raising the
effective rank of a feature Gram matrix is not a precondition for the
$\log\mathrm{PR}$ term: it is a property that can be pushed on the features of any
classifier, independent of how that classifier was trained. It is instead a
post-hoc account of why this term is particularly effective
against \CondNum{} specifically. \CondNum{}'s own training objective minimizes
\begin{equation}
  R_{\mathrm{ill}}(G) = \Bigl[\tfrac{1}{2K}\lVert G \rVert_F^2 -
  \tfrac{1}{2}\bigl(\sigma_{\min}^G\bigr)^2\Bigr]^{-1} ,
  \label{eq:a3-cn-rill}
\end{equation}
through the weights during immunization, where $K$ is the dimension of $G$, and the attacker's $\log\mathrm{PR}$ term is
the scale-invariant form of this same quantity, so maximizing it through the
input directly reverses what \CondNum{} minimizes through the weights.

\paragraph{Model-aware perturbation for generation.}
PGA maximizes $\mathcal{L}_{\text{diff}}(M_{\text{immu}};x+\delta)$ by iterating the projected step
$\delta \leftarrow \Pi_{\mathcal{C}_{\mathrm{pix}}}\bigl(\delta +
\alpha\,\mathrm{sign}(\nabla_\delta \mathcal{L}_{\text{diff}})\bigr)$ for $40$ steps, where $\Pi_{\mathcal{C}_{\mathrm{pix}}}$ projects onto the pixel budget
$\mathcal{C}_{\mathrm{pix}}=\{\delta:\lVert\delta\rVert_\infty\leq\varepsilon\}$ with $\varepsilon=4/255$, with step size $\alpha=0.4/255$ and a random
start. We take the expectation over diffusion timesteps and horizontal flips
to match the attacker's training transformations. Because the loss is
evaluated on the immunized weights, \IMMA{} and CLAMP receive separate
perturbations. Each defense also receives a model-agnostic control at the same budget.

\paragraph{Combined attack on generation.}
The Combined attack applies PGA to the SDXL images and uses the corresponding
name-free style description. No other adaptation or evaluation setting
changes.

\paragraph{Model-aware perturbation for language models.}
GCG optimizes a $20$-token suffix on the immunized model using cross-entropy
toward the harmful target answer plus a refusal suppression term with weight
$\lambda=0.5$. This weight is calibrated by direct search: $\lambda=2$ forces
the model into incoherent output, while $\lambda=0.5$ yields coherent harmful
content. We use $250$ steps, top $k=256$, and $512$ candidates per step.

The suffix is then appended to each of the $238$ attack requests, each paired
with its harmful answer, and the attacker adapts the immunized model on
that set: LoRA of rank $r=8$ and $\gamma=2$ on the attention $q$ and $v$
projections, where $\gamma$ is the LoRA alpha, so the adapter update is scaled
by $\gamma/r$, learning rate $10^{-4}$, five epochs, effective batch $8$.
Reported harmful rate is the mean over epochs three through five on the $358$
held-out requests.

The matched control replaces the optimized suffix with the best of $N$ random
ones and changes nothing else. We draw $N=128{,}000$ suffixes uniformly from
the vocabulary GCG may use, the whole candidate budget GCG spends over its
$250$ steps, and score them with GCG's own objective, imported from the GCG
implementation. Scoring runs in two
stages: all $N$ candidates on one fixed batch of $8$ requests, then the best
$256$ rescored on all $238$. No gradient is used at any point.

\section{Evaluation metrics and controls}
\label{sec:threat:criteria}

This section states the reporting conventions and controls that the metric
definitions of Sec.~\ref{sec:exp:metrics} leave implicit.

\paragraph{Reporting.}
On classification, $\mathrm{SGR}$ instantiates Eq.~\ref{eq:sgr} with probe test accuracy at
epochs $10/20/30/40/50$. Tab.~\ref{tab:a3-cn} reports it beside the probe
accuracies averaged over the same epochs. On generation, the relative gap is computed
separately for CLIP, DINO, and $1-\mathrm{LPIPS}$ at LoRA checkpoints
$10/20/30/40/50$, averaged over checkpoints, and then averaged over the three
metrics. In every modality, $\mathrm{SGR}$ is a mean of ratios: the ratio of
Eq.~\ref{eq:sgr} is taken per checkpoint and then averaged, so it
is not the ratio of the averaged utilities printed beside it.

\paragraph{Controls.}
The model-aware pixel perturbations use PGA under an $\ell_\infty$
budget~\citep{pgd2018}, with expectation over transformation~\citep{eot2018}
for the attacker's sampling randomness. We use this \emph{optimized}
perturbation as an instrument, comparing it with the random
control of the same budget. Such controls follow a standing judgment in adversarial machine
learning: a defense's reported strength is a function of the attack it was
evaluated against. That judgment carries extra weight in the present setting
because the evaluated and the optimized-against attackers coincide by
construction, so we hold the following checks fixed throughout. Model-aware PGA is
compared with model-agnostic noise under the same $\ell_\infty$ budget. Absolute utility is reported beside
every ratio. For classification, $\mathrm{ACC}_O$ checks whether the same
images also damage the original model. For generation, model-agnostic noise and PGA are
compared within each defense because the PGA objective is defense specific.
Together these checks separate a reduction in $\mathrm{SGR}$ that reflects
a gain confined to the immunized arm from one that reflects damage to both
arms alike.
App.~\ref{sec:disc:method} states what these checks establish. All $\pm$ values denote standard deviations over random seeds.

\section{Instantiation details}
\label{app:instantiation}

This appendix instantiates the common threat model for classification, text-to-image
generation, and language models, and Tab.~\ref{tab:threat-instantiation}
summarizes the three instantiations side by side. Every comparison
includes the corresponding plain-adaptation result under the same implementation
and evaluation protocol.

\begin{table*}[t]
\centering
\caption{Instantiation of the threat model of Sec.~\ref{sec:threat} across
the three modalities. All three share the release-time structure and differ in
the adaptation strategy, attack set, and utility.}
\label{tab:threat-instantiation}
\small
\setlength{\tabcolsep}{3pt}
\renewcommand{\arraystretch}{1.08}
\begin{tabular}{
@{}
>{\raggedright\arraybackslash}p{0.15\textwidth}
>{\raggedright\arraybackslash}p{0.22\textwidth}
>{\raggedright\arraybackslash}p{0.29\textwidth}
>{\raggedright\arraybackslash}p{0.27\textwidth}
@{}
}
\toprule
&
\textbf{Diffusion (\IMMA{}/CLAMP)}
&
\textbf{Classification (\CondNum{}/CLAMP)}
&
\textbf{LLM (CLAMP)}
\\
\midrule

Released model $M_{\text{immu}}$
&
Immunized SD-v1.4 U-Net
&
Immunized ResNet-18
&
Immunized instruct LLM
\\

Adaptation $\mathcal{A}$
&
Rank-$4$ LoRA fine-tuning
&
Linear probe, frozen backbone
&
Rank-$8$ LoRA fine-tuning
\\

Attack set $\mathcal{D}_{\text{harm}}^{\text{attk}}$
&
$20$ artist images and captions
&
Stanford Cars, Food-101, Flowers-102, VMMRdb-196
&
$238$ SG-Bench requests with harmful answers
\\

Clean evaluation $R$
&
Reference images
&
Clean test splits
&
$358$ held-out SG-Bench requests
\\

Utility $U$
&
CLIP, DINO, and $1-\mathrm{LPIPS}$
&
Top-$1$ accuracy
&
Harmful rate via Llama-Guard-3-1B and WildGuard
\\

Model-aware budget
&
$\lVert\delta\rVert_\infty\leq4/255$
&
$\lVert\delta\rVert_\infty\leq1/255$ (\CondNum{}), $2/255$ (CLAMP)
&
$20$ tokens
\\

\bottomrule
\end{tabular}
\end{table*}

\paragraph{Text-to-image generation.}
We evaluate \IMMA{} and CLAMP on Van~Gogh, Tyler~Edlin, Kilian~Eng,
Thomas~Kinkade, and Kelly~McKernan. The attacker fine-tunes the immunized U-Net
with a rank-$4$ LoRA adapter on $20$ images and evaluates checkpoints
$10/20/30/40/50$. Unless an attack changes the image source or caption, the
prompt is \texttt{An artwork by <Artist>}.

\begin{figure*}[t]
\centering
\includegraphics[width=\textwidth]{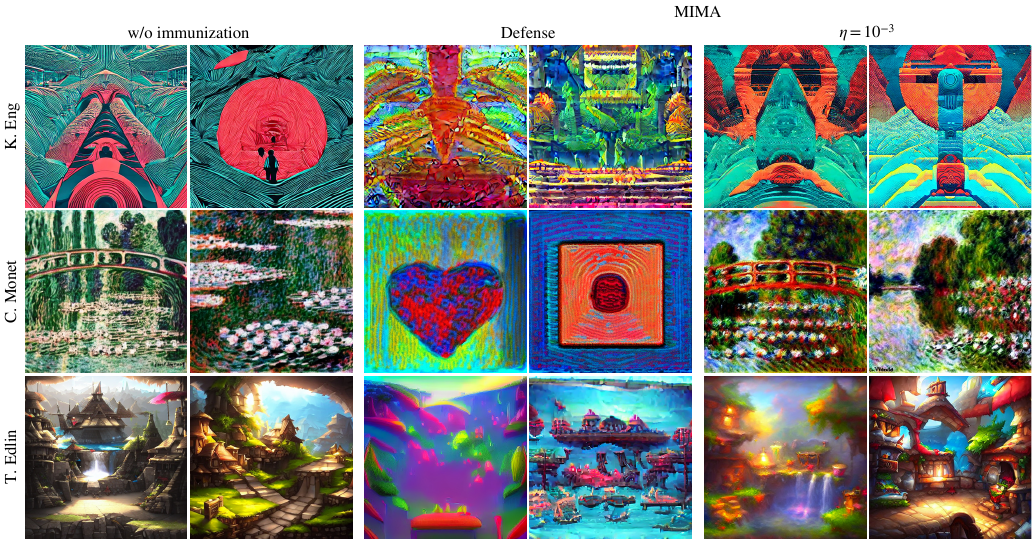}
\caption{Adapted samples for the three artists of Group~3 under MIMA and adaptation misalignment with learning
rate $\eta=10^{-3}$. \emph{w/o immunization} and \emph{Defense} use plain adaptation. The \emph{Defense}
samples lose the artists' styles, and the $\eta=10^{-3}$ samples bring them back.}
\label{fig:mima-samples}
\end{figure*}

\paragraph{Multi-concept immunization (MIMA).}
\label{app:mima}
MIMA~\citep{mima2025} immunizes several concepts at once, and we take each artist
style as one concept. For each group of artists, we erase the group's styles with
UCE~\citep{uce2024} to obtain $M_{\text{prim}}$ and then run the official MIMA
implementation to obtain $M_{\text{immu}}$. The attacker fine-tunes one rank-$4$ LoRA
adapter on the images of all artists of the group, using the prompt
\texttt{An artwork of <Artist>}, and $\mathrm{SGR}$ is averaged over the three
metrics and the artists of the group. Prompt misalignment replaces each caption with
the name-free description of Tab.~\ref{tab:a2-prompts}, and the learning-rate change
sets $\eta$ to $3\times10^{-4}$ or $10^{-3}$. $M_{\text{prim}}$ is adapted under the
same change when measuring $U_O$. Tab.~\ref{tab:mima-full} adds $\eta=3\times10^{-4}$
and the change in style similarity relative to plain adaptation. $\mathrm{SGR}$ is
lower than under plain adaptation in every cell. $\Delta U_I$ rises in every cell, so the attacker obtains a higher similarity from
$M_{\text{immu}}$. At $\eta=10^{-3}$, $U_I$ exceeds $U_O$ for every artist of every group, by between
$+0.007$ and $+0.066$.
Fig.~\ref{fig:mima-samples} shows adapted samples for Group~3 at $\eta=10^{-3}$.

\begin{table*}[t]
\centering
\caption{MIMA under prompt misalignment and a changed learning rate $\eta$, for the groups of
Tab.~\ref{tab:mima}. $\Delta U_O$ and $\Delta U_I$ are the changes in style similarity from plain
adaptation after adapting $M_{\text{prim}}$ and $M_{\text{immu}}$.}
\label{tab:mima-full}
\small
\renewcommand{\arraystretch}{1.15}
\setlength{\tabcolsep}{3.3pt}
\begin{tabular}{l cccc ccc ccc}
\toprule
 & \multicolumn{4}{c}{$\mathrm{SGR}$} & \multicolumn{3}{c}{$\Delta U_O$} & \multicolumn{3}{c}{$\Delta U_I$} \\
\cmidrule(lr){2-5} \cmidrule(lr){6-8} \cmidrule(lr){9-11}
 & Plain adapt. & Prompt & $\eta{=}3{\times}10^{-4}$ & $\eta{=}10^{-3}$ & Prompt & $\eta{=}3{\times}10^{-4}$ & $\eta{=}10^{-3}$ & Prompt & $\eta{=}3{\times}10^{-4}$ & $\eta{=}10^{-3}$ \\
\midrule
Group 1 & $25.60_{\pm 8.24}$ & $12.17_{\pm 2.21}$ & $5.65_{\pm 3.76}$ & $-5.07_{\pm 4.11}$ & $+0.012$ & $-0.004$ & $-0.031$ & $+0.101$ & $+0.121$ & $+0.157$ \\
Group 2 & $16.61_{\pm 6.62}$ & $9.36_{\pm 2.38}$ & $-1.97_{\pm 4.84}$ & $-7.06_{\pm 4.76}$ & $+0.016$ & $-0.014$ & $-0.039$ & $+0.057$ & $+0.095$ & $+0.097$ \\
Group 3 & $15.61_{\pm 3.22}$ & $5.17_{\pm 1.94}$ & $7.69_{\pm 8.22}$ & $-6.82_{\pm 2.94}$ & $+0.032$ & $-0.001$ & $-0.028$ & $+0.088$ & $+0.042$ & $+0.091$ \\
Group 4 & $25.27_{\pm 5.73}$ & $6.09_{\pm 1.84}$ & $6.01_{\pm 5.42}$ & $-2.25_{\pm 3.88}$ & $+0.007$ & $-0.007$ & $-0.023$ & $+0.125$ & $+0.111$ & $+0.144$ \\
\bottomrule
\end{tabular}
\end{table*}

\begin{table*}[t]
\centering
\caption{Name-free style descriptions used by the prompt misalignment attack.}
\label{tab:a2-prompts}
\small
\begin{tabular}{p{0.15\textwidth}p{0.78\textwidth}}
\toprule
Artist & Name-free prompt \\
\midrule
Van Gogh &
a post-impressionist oil painting, swirling brushstrokes, thick impasto,
vivid blue and yellow \\
\addlinespace
Tyler Edlin &
an epic fantasy environment concept art, dramatic painterly landscape,
cinematic lighting \\
\addlinespace
Kilian Eng &
a psychedelic retro-futuristic sci-fi illustration, bold flat color planes in
orange teal and purple, intricate fine linework, surreal alien dreamscape,
vintage 1970s science-fiction comic art \\
\addlinespace
Thomas Kinkade &
a luminous idyllic cottage painting, glowing golden window light at dusk,
blossoming gardens along a winding stone path, warm amber and rose highlights,
nostalgic romantic realism \\
\addlinespace
Kelly McKernan &
an ethereal watercolor and gouache figure illustration, translucent bleeding
washes and delicate ink linework, iridescent teal magenta and violet, flowing
celestial and floral motifs, art nouveau elegance \\
\addlinespace
Claude Monet &
an impressionist plein-air oil painting, loose broken brushstrokes, shimmering light
on water and garden flowers, soft pastel blue green and lilac, hazy atmospheric
reflections \\
\bottomrule
\end{tabular}
\end{table*}

\paragraph{Classification feature extractor.}
\CondNum{} and CLAMP each release an immunized ResNet-18 feature extractor. The attacker freezes
the backbone and trains a linear probe for $50$ epochs on its $512$-dimensional
features. We evaluate Stanford Cars, Food-101, Flowers-102,
and VMMRdb-196 on their clean test sets.

\paragraph{Language models.}
We immunize an already safety-aligned instruction model with CLAMP, keeping
its base weights frozen. SG-Bench is split into $596$ defense examples, $238$ attack examples,
and $358$ held-out evaluation examples. The attacker adapts the immunized
model on the $238$ attack examples paired with their harmful answers,
and the three evaluated models answer the held-out requests through greedy
free generation, with Llama-Guard-3-1B and WildGuard~\citep{wildguard2024}
separately grading each response as safe or unsafe.

\paragraph{Classification models.}
\label{app:cls-ggp}
The attacker's own adaptation strategy fits a linear probe on the features the frozen, released backbone produces for its harmful data, so the probe can only separate classes along whatever directions those features actually occupy. If the released backbone maps the attacker's harmful data into few effective directions, the probe has little signal to work with no matter how it is trained. The attacker therefore perturbs its own inputs with two goals: the probe trained on the perturbed features should reach a high accuracy on clean images, and the released backbone should spread those features across more directions before the probe ever sees them.
The attacker perturbs its inputs with $\delta$, subject to the pixel budget
$\mathcal{C}_{\mathrm{pix}}$ of App.~\ref{app:method}, and solves
\begin{equation}
  \delta^{*}=\arg\max_{\delta\in\mathcal{C}_{\mathrm{pix}}}\,\widehat{\mathrm{Acc}}\bigl(W(\delta)\bigr)+\beta\log \mathrm{PR}\bigl(\Phi_{M_{\text{immu}}}(x+\delta)\bigr) ,
  \label{eq:a3-cn-obj}
\end{equation}
where $\Phi_{M_{\text{immu}}}$ is the frozen feature extractor of the released model,
$W(\delta)$ is the linear probe obtained by training on the perturbed features
$\Phi_{M_{\text{immu}}}(x+\delta)$, $\widehat{\mathrm{Acc}}$ is a differentiable
accuracy of that probe on clean images, $\mathrm{PR}$ is the participation ratio, and
$\beta=1$ weights the two terms. The first term simulates the accuracy that the
attacker's probe will reach. The second term spreads the features, which immunization
concentrates in a few directions, back over many directions.

\textit{Simulated probe training.}
Training the probe for $50$ epochs inside every PGA step leaves no tractable gradient
with respect to $\delta$, so $W(\delta)$ is computed by a closed-form surrogate of that
training in which every step is differentiable. Let
$A_\delta\in\mathbb{R}^{n\times(d+1)}$ stack the perturbed features of the $n$ fitting
images, each extended by a constant $1$ for the bias, with $d=512$, and let
$Y\in\{0,1\}^{n\times C}$ hold their one-hot labels over $C$ classes. The surrogate
depends on $\delta$ only through
\begin{equation}
  M_\delta=\tfrac{1}{n}A_\delta^{\top}A_\delta ,\qquad
  S_\delta=\tfrac{1}{n}\bigl(Y-\tfrac{1}{C}\mathbf{1}_n\mathbf{1}_C^{\top}\bigr)^{\!\top}A_\delta ,
  \label{eq:a3-cf-stats}
\end{equation}
the uncentered second moment of the features and their class signal. Linearizing the
softmax cross-entropy around the uniform prediction, with its curvature replaced by the
constant $\kappa/C$, turns the probe gradient into
$\tfrac{\kappa}{C}PWM_\delta-S_\delta$, where $P=I_C-\tfrac{1}{C}\mathbf{1}_C\mathbf{1}_C^{\top}$
subtracts the class mean from each column of the probe weights
$W\in\mathbb{R}^{C\times(d+1)}$, and $\mathbf{1}_k$ is the all-ones vector of length $k$. The surrogate unrolls SGD on this gradient with the
momentum $\mu$, weight decay $\lambda$, and learning rate $\eta_t$ of the attacker's
probe training,
\begin{equation}
\begin{aligned}
  V_{t+1}&=\mu V_t+\tfrac{\kappa}{C}PW_tM_\delta-S_\delta+\lambda W_t ,\\
  W_{t+1}&=W_t-\eta_t V_{t+1} ,
\end{aligned}
  \label{eq:a3-cf-sgd}
\end{equation}
from $V_0=0$ and $W_0$ drawn from the default initialization of a linear layer, and
keeps $W_t$ at the steps that correspond to epochs $10/20/30/40/50$, the checkpoints at
which $\mathrm{SGR}$ is evaluated.

\textit{Clean, held-out score.}
To keep the perturbation from writing the labels into the images, the score is
cross-fitted. The training images are split into five folds, $W(\delta)$ is fit on the
perturbed features of four folds, and it is scored on the clean features of the fifth,
which were neither perturbed nor used in the fit. The score is a soft accuracy: for the
held-out fold $H$, with clean bias-extended features $\tilde a_j$, labels $y_j$, and
logits $z_j=W\tilde a_j$,
\begin{equation}
  \widehat{\mathrm{Acc}}(W)=\frac{1}{|H|}\sum_{j\in H}\mathrm{softmax}\bigl(z_j/T\bigr)_{y_j} ,
  \label{eq:a3-cf-acc}
\end{equation}
where the temperature $T$ is $0.25$ times the standard deviation of the logits. The
score is averaged over the five checkpoints and the five folds, and against CLAMP
also over the augmented views of App.~\ref{app:method}.

\textit{Feature spread.}
The second term of Eq.~\ref{eq:a3-cn-obj} uses
\begin{equation}
  \mathrm{PR}(G) = \frac{(\operatorname{tr} G)^2}{\lVert G \rVert_F^2}, \quad G = \Phi_{M_{\text{immu}}}(x+\delta)^{\!\top}\Phi_{M_{\text{immu}}}(x+\delta),
  \label{eq:a3-cn-pr}
\end{equation}
a differentiable stand-in for the effective rank of the feature Gram matrix $G$, a
property any feature representation has regardless of how it was produced: energy
spread evenly over $k$ directions gives $\mathrm{PR}=k$, energy sitting entirely in one
direction gives $\mathrm{PR}=1$, and both extremes need only the trace and the Frobenius
norm, without an eigendecomposition. The attacker maximizes this quantity through the
input.

\textit{Balancing the two terms.}
The gradients of the two terms differ widely in scale, so adding them directly
lets one term drown out the other. We therefore apply the weight $\beta$ to normalized
gradients. With $g_{\mathrm{acc}}$ and $g_{\mathrm{PR}}$ the gradients of the two terms
of Eq.~\ref{eq:a3-cn-obj} with respect to the features $\Phi_{M_{\text{immu}}}(x+\delta)$,
we combine them as
\begin{equation}
  g_{\Phi}=g_{\mathrm{acc}}+\beta\,\frac{\lVert g_{\mathrm{acc}}\rVert_1}{\lVert g_{\mathrm{PR}}\rVert_1}\,g_{\mathrm{PR}} ,
  \label{eq:a3-cf-vote}
\end{equation}
and backpropagate $g_{\Phi}$ through $\Phi_{M_{\text{immu}}}$ to obtain the ascent
direction $g$ on $\delta$, so with $\beta=1$ the two terms carry equal $\ell_1$ weight
in every step.

See App.~\ref{app:method} for implementation-level detail on the pixel
perturbation above.

\paragraph{Language-model suffix.}
The immunized model answers a harmful request with an
answer that opens with a refusal, which suppresses the likelihood it ever continues into
the harmful content the request asked for. The suffix reverses both halves, lifting the
harmful answer's likelihood while suppressing the refusal onset.
The attacker appends a single suffix $s$ of fixed length to each of its harmful requests,
pairing every suffixed request $q_i\oplus s$ with its harmful answer $a_i$. It optimizes
$s$ against $M_{\text{immu}}$ and then adapts $M_{\text{immu}}$ on the resulting pairs, much as
model-aware PGA perturbs the images it goes on to adapt on. The search space is the
model's discrete vocabulary raised to the suffix length, excluding every special and
control token, such as an end-of-sequence or unused-slot marker. The suffix
minimizes the objective of App.~\ref{app:llm-method}, averaged over the batch:
\begin{equation}
  s^{*}=\arg\min_{s\in\mathcal{V}^{L}}\bigl[-\lambda\log(1-P_{\text{refusal}})-\log P_{M_{\text{immu}}}(a\mid q\oplus s)\bigr] ,
  \label{eq:a3-llm-obj}
\end{equation}
where $\log P_{M_{\text{immu}}}(a\mid q\oplus s)$ is averaged over the tokens of the harmful
answer $a$, truncated to a fixed number of tokens, so a lower objective means the
model is more willing to produce that content. Whether the model would ultimately refuse is only settled once a full response has
been sampled, which the search cannot afford at every candidate, so
$P_{\text{refusal}}$ scores refusal at the point where it is already decided: it is the
probability, at the first generated token, of landing on any of a fixed set of
refusal-onset tokens, such as ``I'm sorry'', ``I cannot'', ``Sorry''; the first term
suppresses it, and only this loss carries a refusal term because a language model,
unlike the other modalities, can decline to answer and still be judged as following the
prompt. See App.~\ref{app:method} for the suffix length, the value of $\lambda$, and how
it is calibrated.
The suffix is found by greedy coordinate-gradient search~\citep[GCG;][]{gcg2023}, the
discrete case of the gradient-guided search described above.
\section{Image misalignment details}
\label{app:a1}

This appendix documents how the substitute fine-tuning images used by the
image misalignment attack in Sec.~\ref{sec:exp:diffusion} are
produced. It also reports a related analysis on classification, in which the
defender's own training images are synthetic.

\paragraph{Protocol.}
The defender immunizes against a style as rendered by Stable Diffusion v1.4.
For each artist, the attacker instead samples $20$ images from Stable
Diffusion XL (SDXL)~\citep{sdxl2024}, a publicly released checkpoint the defender never
immunized against, using the defender's own prompt \texttt{An artwork by <Artist>}.
Those $20$ images are the LoRA fine-tuning set. The immunized model, the
rank-$4$ adapter, the optimization schedule, and the prompt are unchanged, so
the checkpoint that produced the attacker's training images is the only
variable in the comparison. No pixel perturbation is applied in this
experiment.

\paragraph{Sampling configuration.}
The images are sampled with $30$ denoising steps and a guidance scale of $7.5$.
Each image is generated at $1024\times1024$ and resized to $512\times512$, so
that resolution does not become a second variable.

\begin{table}[t]
\centering
\caption{Defense-training distribution generalization on \CondNum{}. \emph{Real} immunizes on the
real training set; \emph{FLUX} immunizes on FLUX.2 synthetic images.
The attacker probes on the real training set and every row is scored on the
real test set. Accuracy is measured at epoch $50$, while $\mathrm{SGR}$
averages epochs $10/20/30/40/50$.}
\label{tab:a1-cls}
\footnotesize
\setlength{\tabcolsep}{3pt}
\begin{tabular}{llccc}
\toprule
Dataset & $\mathcal{D}_{\text{harm}}^{\text{immu}}$ & $\mathrm{ACC}_O@50$ & $\mathrm{ACC}_I@50$ & $\mathrm{SGR}$ \\
\midrule
\multirow{2}{*}{Stanford Cars} & Real & \multirow{2}{*}{$32.95_{\pm 0.37}$} & $28.08_{\pm 1.85}$ & $20.99_{\pm 5.90}$ \\
& FLUX & & $33.04_{\pm 0.73}$ & $6.71_{\pm 1.26}$ \\
\addlinespace
\multirow{2}{*}{Flowers-102} & Real & \multirow{2}{*}{$78.11_{\pm 0.33}$} & $78.73_{\pm 0.68}$ & $5.61_{\pm 0.99}$ \\
& FLUX & & $79.91_{\pm 0.16}$ & $-4.98_{\pm 0.61}$ \\
\addlinespace
\multirow{2}{*}{VMMRdb-196} & Real & \multirow{2}{*}{$14.70_{\pm 0.61}$} & $10.81_{\pm 0.80}$ & $35.58_{\pm 3.01}$ \\
& FLUX & & $15.31_{\pm 0.63}$ & $6.45_{\pm 1.79}$ \\
\bottomrule
\end{tabular}
\vspace{-10pt}
\end{table}

\begin{table*}[t]
\centering
\caption{Prompts used by the adaptive defender. Each prompt labels four of the defender's twenty images.}
\label{tab:adaptive-prompts}
\small
\begin{tabular}{p{0.15\textwidth}>{\raggedright\arraybackslash}p{0.78\textwidth}}
\toprule
Artist & Prompts in the immunization set \\
\midrule
Van Gogh & An artwork by Van Gogh; A painting by Van Gogh; An oil painting by Van Gogh; A canvas artwork in the style of Van Gogh; A fine art painting by Van Gogh \\
\addlinespace
Tyler Edlin & An artwork by Tyler Edlin; A painting by Tyler Edlin; A digital artwork by Tyler Edlin; A fantasy environment artwork by Tyler Edlin; A concept art piece by Tyler Edlin \\
\addlinespace
Kilian Eng & An artwork by Kilian Eng; A painting by Kilian Eng; An illustration by Kilian Eng; A science fiction artwork by Kilian Eng; A retro futuristic illustration by Kilian Eng \\
\addlinespace
Thomas Kinkade & An artwork by Thomas Kinkade; A painting by Thomas Kinkade; An oil painting by Thomas Kinkade; A landscape painting by Thomas Kinkade; A village scene painted by Thomas Kinkade \\
\addlinespace
Kelly McKernan & An artwork by Kelly McKernan; A painting by Kelly McKernan; An illustration by Kelly McKernan; A fantasy illustration by Kelly McKernan; A whimsical painting by Kelly McKernan \\
\bottomrule
\end{tabular}
\end{table*}

\paragraph{Evaluation reference.}
Similarity is scored against the same $100$ Stable Diffusion v1.4 reference
images for that artist, whichever attack produced the fine-tuning set.
Substituting the generator therefore changes what the attacker learns from and
not what the measurement compares against. CLIP, DINO, and
$1-\mathrm{LPIPS}$ similarities at LoRA checkpoints
$10/20/30/40/50$ are combined into $\mathrm{SGR}$ as described in
App.~\ref{sec:threat:criteria}. Tab.~\ref{tab:gen-main} reports the quantitative results
once; this appendix does not duplicate that table.

\paragraph{Defense-training distribution generalization on classification.}
In practice the defender does not have access to the real harmful training set and instead
approximates $\mathcal{D}_{\text{harm}}^{\text{immu}}$ from the class names alone, using images
synthesized by FLUX.2 [klein] 9B~\citep{flux2klein2026}. On \CondNum{}, this analysis changes only the defender's training
data. The defender generates $\mathcal{D}_{\text{harm}}^{\text{immu}}$
this way and runs \CondNum{} on it, keeping ImageNet~\citep{imagenet2009} as the benign set. The attacker probes the
immunized model on the real training set exactly as in App.~\ref{app:instantiation}. The reference
row is immunized on real images. The original extractor $M_{\text{prim}}$, the attacker's data, and
the real test set are shared, so the data the defender immunized on is the only variable.
Tab.~\ref{tab:a1-cls} reports that the synthetic $\mathcal{D}_{\text{harm}}^{\text{immu}}$ lowers
$\mathrm{SGR}$ on all three datasets, and lowers it below zero on Flowers-102.

\section{Prompt misalignment details}
\label{app:a2}
\label{app:prompt-details}

This appendix provides the exact captions and the supporting weight-space
analysis for the prompt misalignment attack in Sec.~\ref{sec:exp:diffusion}.

\paragraph{Prompt construction.}
Each caption describes the protected visual style without using the artist's
name or another proper noun. Tab.~\ref{tab:a2-prompts} lists them. The image files and immunized model remain
unchanged.

\paragraph{Prompts of the adaptive defender.}
\label{app:adaptive-prompts}
For the adaptive defender in Sec.~\ref{sec:exp:ablation}, \IMMA{} labels its twenty
images with the five prompts in Tab.~\ref{tab:adaptive-prompts}, four images each. All
other immunization settings are unchanged.

\paragraph{Weight space evidence.}
We measure how strongly the immunization displacement amplifies text
directions in the cross attention key and value projections. Across the five
concepts, the protected artist's own name is amplified $58$ to $111$ times,
whereas other artists' names in the same template and token positions are
amplified only $11$ to $15$ times. Random directions remain close to one. This
concentration on the protected name supports the behavioral result: a name-free
description can preserve the visual target while avoiding the strongest
textual direction introduced by immunization.

\section{Model-aware perturbation details}
\label{app:a3}

This appendix gives further classification and generation results behind
Sec.~\ref{sec:exp:classification} and Sec.~\ref{sec:exp:diffusion}.

\begin{table}[t]
\centering
\caption{Participation ratio of the features of the attacker's training images
under \CondNum{}. Noise and PGA are the perturbations of Tab.~\ref{tab:a3-cn}.
Bold marks the highest $\mathrm{PR}$ among the $M_{\text{immu}}$ columns.}
\label{tab:rill}
\small
\setlength{\tabcolsep}{6pt}
\renewcommand{\arraystretch}{1.15}
\begin{tabular}{lcccc}
\toprule
\multirow{2}{*}{Dataset} & \multirow{2}{*}{$M_{\text{prim}}$} & \multicolumn{3}{c}{$M_{\text{immu}}$} \\
\cmidrule(lr){3-5}
 & & Clean & Noise & PGA \\
\midrule
Stanford Cars & $2.02$ & $1.49$ & $2.03$ & $\mathbf{3.20}$ \\
Food-101      & $2.47$ & $1.81$ & $1.81$ & $\mathbf{3.23}$ \\
Flowers-102   & $2.26$ & $1.39$ & $1.39$ & $\mathbf{2.11}$ \\
VMMRdb-196    & $1.76$ & $1.23$ & $1.23$ & $\mathbf{1.48}$ \\
\bottomrule
\end{tabular}
\end{table}

\begin{table*}[t]
\centering
\caption{$\mathrm{ACC}_O$ (\%) under the two attacks against \CondNum{} and CLAMP, averaged over
epochs $10/20/30/40/50$.}
\label{tab:attk-acco}
\vspace{3pt}
\small
\setlength{\tabcolsep}{4pt}
\begin{tabular}{lcccc}
\toprule
& \multicolumn{2}{c}{\CondNum{}} & \multicolumn{2}{c}{CLAMP} \\
\cmidrule(lr){2-3}\cmidrule(lr){4-5}
Dataset & Model-agnostic & Model-aware & Model-agnostic & Model-aware \\
\midrule
Stanford Cars & $28.04_{\pm 0.06}$ & $27.80_{\pm 0.12}$ & $15.47_{\pm 0.12}$ & $16.02_{\pm 0.11}$ \\
Food-101 & $47.66_{\pm 0.07}$ & $47.77_{\pm 0.17}$ & $29.71_{\pm 0.13}$ & $34.15_{\pm 0.28}$ \\
Flowers-102 & $75.44_{\pm 0.30}$ & $74.92_{\pm 0.16}$ & $41.78_{\pm 0.46}$ & $42.17_{\pm 0.44}$ \\
VMMRdb-196 & $11.59_{\pm 0.40}$ & $12.10_{\pm 0.29}$ & $5.11_{\pm 0.34}$ & $6.08_{\pm 0.42}$ \\
\bottomrule
\end{tabular}
\end{table*}

\begin{table*}[t]
\centering
\caption{Absolute style similarity behind the generation results. $U_O$ and
$U_I$ are the similarities reached after adapting $M_{\text{prim}}$ and
$M_{\text{immu}}$, averaged over checkpoints, and $\mathrm{SGR}$ is computed per checkpoint as in
App.~\ref{sec:threat:criteria}.}
\label{tab:gen-full}
\small
\begin{tabular}{lllcccccc}
\toprule
& & & \multicolumn{3}{c}{Plain adaptation} & \multicolumn{3}{c}{Model-aware (PGA)} \\
\cmidrule(lr){4-6}\cmidrule(lr){7-9}
Defense & Artist & Metric & $U_O$ & $U_I$ & SGR & $U_O$ & $U_I$ & SGR \\
\midrule
\multirow{15}{*}{\IMMA{}} & \multirow{3}{*}{Van~Gogh} & CLIP & $0.8298$ & $0.8002$ & $3.58$ & $0.8301$ & $0.8115$ & $2.21$\\
 &  & DINO & $0.4883$ & $0.3521$ & $27.88$ & $0.4845$ & $0.4205$ & $13.23$\\
 &  & $1-$LPIPS & $0.6292$ & $0.4946$ & $21.42$ & $0.6377$ & $0.5746$ & $9.80$\\
\cmidrule(l){2-9}
 & \multirow{3}{*}{T.~Edlin} & CLIP & $0.9093$ & $0.8097$ & $10.94$ & $0.9101$ & $0.8397$ & $7.75$\\
 &  & DINO & $0.6183$ & $0.4248$ & $31.47$ & $0.6144$ & $0.4705$ & $23.63$\\
 &  & $1-$LPIPS & $0.6968$ & $0.6030$ & $13.46$ & $0.6921$ & $0.6202$ & $10.43$\\
\cmidrule(l){2-9}
 & \multirow{3}{*}{K.~Eng} & CLIP & $0.9061$ & $0.8307$ & $8.32$ & $0.8981$ & $0.8539$ & $4.91$\\
 &  & DINO & $0.5322$ & $0.3345$ & $37.35$ & $0.5173$ & $0.4086$ & $21.85$\\
 &  & $1-$LPIPS & $0.6607$ & $0.5309$ & $19.72$ & $0.6531$ & $0.6094$ & $6.70$\\
\cmidrule(l){2-9}
 & \multirow{3}{*}{T.~Kinkade} & CLIP & $0.8974$ & $0.8232$ & $8.31$ & $0.8795$ & $0.8302$ & $5.62$\\
 &  & DINO & $0.6281$ & $0.4509$ & $28.40$ & $0.5986$ & $0.4756$ & $21.50$\\
 &  & $1-$LPIPS & $0.6877$ & $0.5603$ & $18.56$ & $0.6732$ & $0.5951$ & $12.09$\\
\cmidrule(l){2-9}
 & \multirow{3}{*}{K.~McKernan} & CLIP & $0.8989$ & $0.7864$ & $12.53$ & $0.8734$ & $0.7935$ & $9.19$\\
 &  & DINO & $0.6175$ & $0.3682$ & $40.78$ & $0.5566$ & $0.4053$ & $28.23$\\
 &  & $1-$LPIPS & $0.6982$ & $0.5770$ & $17.39$ & $0.6808$ & $0.6039$ & $11.38$\\
\midrule
\multirow{15}{*}{CLAMP} & \multirow{3}{*}{Van~Gogh} & CLIP & $0.8298$ & $0.7949$ & $4.20$ & $0.8230$ & $0.7945$ & $3.43$\\
 &  & DINO & $0.4883$ & $0.3750$ & $23.27$ & $0.4832$ & $0.4170$ & $13.61$\\
 &  & $1-$LPIPS & $0.6292$ & $0.4546$ & $27.77$ & $0.6327$ & $0.5480$ & $13.10$\\
\cmidrule(l){2-9}
 & \multirow{3}{*}{T.~Edlin} & CLIP & $0.9093$ & $0.7175$ & $21.09$ & $0.9112$ & $0.7345$ & $19.39$\\
 &  & DINO & $0.6183$ & $0.2472$ & $59.98$ & $0.6236$ & $0.2711$ & $56.52$\\
 &  & $1-$LPIPS & $0.6968$ & $0.4239$ & $39.17$ & $0.7010$ & $0.4961$ & $29.26$\\
\cmidrule(l){2-9}
 & \multirow{3}{*}{K.~Eng} & CLIP & $0.9061$ & $0.8497$ & $6.22$ & $0.9108$ & $0.8662$ & $4.90$\\
 &  & DINO & $0.5322$ & $0.3174$ & $40.65$ & $0.5420$ & $0.3579$ & $34.16$\\
 &  & $1-$LPIPS & $0.6607$ & $0.5458$ & $17.54$ & $0.6650$ & $0.5787$ & $13.07$\\
\cmidrule(l){2-9}
 & \multirow{3}{*}{T.~Kinkade} & CLIP & $0.8974$ & $0.7769$ & $13.45$ & $0.9012$ & $0.8068$ & $10.49$\\
 &  & DINO & $0.6281$ & $0.3302$ & $47.43$ & $0.6195$ & $0.3937$ & $36.49$\\
 &  & $1-$LPIPS & $0.6877$ & $0.4769$ & $30.66$ & $0.6876$ & $0.5629$ & $18.23$\\
\cmidrule(l){2-9}
 & \multirow{3}{*}{K.~McKernan} & CLIP & $0.8989$ & $0.7085$ & $21.20$ & $0.8813$ & $0.7018$ & $20.36$\\
 &  & DINO & $0.6175$ & $0.2448$ & $60.34$ & $0.5626$ & $0.2600$ & $53.76$\\
 &  & $1-$LPIPS & $0.6982$ & $0.4887$ & $30.00$ & $0.6863$ & $0.5063$ & $26.21$\\
\bottomrule
\end{tabular}
\end{table*}

\begin{figure*}[t]
\centering
\includegraphics[width=0.75\textwidth]{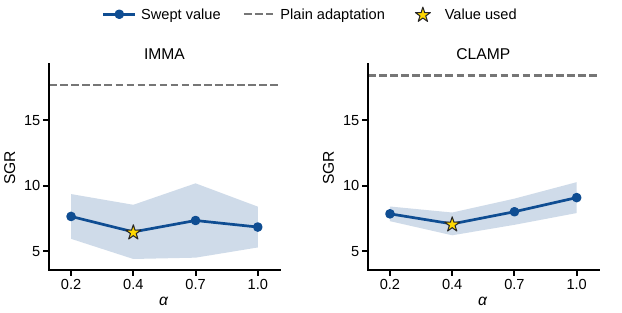}
\caption{Step-size sweeps. Each panel varies the PGA step size $\alpha$, in
units of $1/255$, of the Combined attack on Van Gogh under \IMMA{} (left) and
CLAMP (right) ($\mathrm{SGR}$). Dashed lines mark the plain-adaptation
value, and the shaded band around each swept line is $\pm$ one standard
deviation. Stars mark the value used elsewhere in the paper
($\alpha{=}0.4/255$), which was fixed before this sweep was run; in each
panel the starred value also has the lowest mean SGR of the four swept
values.}
\label{fig:budget-ablation}
\end{figure*}

\paragraph{Feature conditioning under the model-aware perturbation.}
\label{app:rill}
The $\log\mathrm{PR}$ term of the model-aware perturbation for classification
(App.~\ref{app:cls-ggp}) maximizes the scale-invariant participation ratio $\mathrm{PR}$ of
Eq.~\ref{eq:a3-cn-pr} rather than minimizing the condition number directly.
$\mathrm{PR}$ is invariant to rescaling the features, so the optimizer cannot
raise it by merely shrinking feature norms. Tab.~\ref{tab:rill} reports $\mathrm{PR}$
for the \CondNum{} features of the attacker's training images. Immunization lowers
$\mathrm{PR}$ on every dataset. The model-aware perturbation raises it on every
dataset, above the level of $M_{\text{prim}}$ on Stanford Cars and Food-101 and close
to it on Flowers-102. In contrast, model-agnostic noise at the same budget leaves it
unchanged on every dataset except Stanford Cars. VMMRdb-196, where $\mathrm{PR}$
recovers only about half of its drop, is also the dataset on which
Tab.~\ref{tab:a3-cn} shows the most protection remaining under \CondNum{} after the attack.

\paragraph{Original-model utility under the attacks.}
\label{app:attack-o}
Tab.~\ref{tab:attk-acco} reports the utility of
$M_{\text{prim}}$ under each attack, $\mathrm{ACC}_O$ for classification.
App.~\ref{app:llm} reports $\mathrm{HR}_O$ for language models.

\paragraph{Full generation results.}
\label{app:gen-full}
Tab.~\ref{tab:gen-full} reports the absolute style similarity behind the
generation numbers, so that a fall in SGR can be attributed. Across the thirty
defense--artist--metric rows, $U_O$ moves by at most $0.061$ between the two
conditions and by more than $0.01$ in $10$ rows, while $U_I$ rises in
$28$ of them by up to $0.093$, about $3.1$ times the movement of $U_O$ on
average. The attacker therefore obtains a higher similarity from the immunized model in
nearly every row, so the lower SGR reflects a gain for the attacker and not only a
smaller ratio.
Tab.~\ref{tab:gen-du} summarizes this change per defense and artist, averaged over
the three metrics.

\begin{table*}[t]
\centering
\caption{Change in style similarity from plain adaptation to model-aware PGA,
averaged over the three metrics. $\Delta U_O$ and $\Delta U_I$ are the changes
in the similarity reached after adapting $M_{\text{prim}}$ and
$M_{\text{immu}}$.}
\label{tab:gen-du}
\footnotesize
\renewcommand{\arraystretch}{1.15}
\setlength{\tabcolsep}{1.5pt}
\begin{tabular}{llccccc}
\toprule
Defense & & Van Gogh & T. Edlin & K. Eng & T. Kinkade & K. McKernan \\
\midrule
\IMMA{} & $\Delta U_O$ & $+0.002$ & $-0.003$ & $-0.010$ & $-0.021$ & $-0.035$ \\
        & $\Delta U_I$ & $+0.053$ & $+0.031$ & $+0.059$ & $+0.022$ & $+0.024$ \\
\midrule
CLAMP   & $\Delta U_O$ & $-0.003$ & $+0.004$ & $+0.006$ & $-0.002$ & $-0.028$ \\
        & $\Delta U_I$ & $+0.045$ & $+0.038$ & $+0.030$ & $+0.060$ & $+0.009$ \\
\bottomrule
\end{tabular}
\end{table*}

\paragraph{Budget sweeps.}
\label{app:budget-sweep}
Fig.~\ref{fig:budget-ablation} sweeps the
PGA step size $\alpha$ of the Combined attack on generation around the value used elsewhere. That
value was fixed before the sweep was run, and the sweep only checks how
sensitive the attack is to it. Every swept value lowers SGR below the
plain-adaptation baseline, so the attack does not hinge on the particular $\alpha{=}0.4/255$. Differences among the swept values are small relative to the gap to the baseline and largely within one standard deviation of one another, so the attack is insensitive to the step size.

\section{Results on language models}
\label{app:llm}

This appendix examines CLAMP on language models. It gives the
setup, the model-aware perturbation for text, and the results of the GCG attack.

\begin{table*}[t]
\centering
\caption{CLAMP language-model results. Each block scores the same responses with a different judge,
Llama-Guard-3-1B~\citep{llamaguard2023,llama3_2024} or
WildGuard~\citep{wildguard2024}. For $\mathrm{SGR}$, lower is better for
the attacker. $\mathrm{HR}_I$ (\%) is the harmful rate after
adapting $M_{\text{immu}}$. $\Delta\mathrm{HR}_I$ is the relative
change of $\mathrm{HR}_I$ from plain adaptation. The $\mathrm{SGR}$ of each attack uses as
un-immunized reference $M_{\text{prim}}$ adapted on the same suffixed requests.
Bold marks the lowest $\mathrm{SGR}$ of each row. Tab.~\ref{tab:attk-hro} reports
$\mathrm{HR}_O$, the harmful rate after adapting $M_{\text{prim}}$, under the two attacks.}
\label{tab:llm-gcg}
\footnotesize
\setlength{\tabcolsep}{2.5pt}
\resizebox{\textwidth}{!}{%
\begin{tabular}{ll cc ccc ccc}
\toprule
& & \multicolumn{2}{c}{Plain adaptation} & \multicolumn{3}{c}{Random suffix} & \multicolumn{3}{c}{Model-aware (GCG)} \\
\cmidrule(lr){3-4}\cmidrule(lr){5-7}\cmidrule(lr){8-10}
Judge & Model
& $\mathrm{SGR}$ & $\mathrm{HR}_I$
& $\mathrm{SGR}$ & $\mathrm{HR}_I$ & $\Delta\mathrm{HR}_I$
& $\mathrm{SGR}$ & $\mathrm{HR}_I$ & $\Delta\mathrm{HR}_I$ \\
\midrule
\multirow{3}{*}{Llama-Guard-3-1B} & Gemma-2-2B
& $83.79$ & $0.56_{\pm 0.00}$
& $48.71$ & $5.25_{\pm 0.28}$ & $+837.50\%$
& $\mathbf{14.05}$ & $64.74_{\pm 1.59}$ & $+11460.71\%$ \\
 & Qwen2.5-1.5B
& $79.55$ & $1.24_{\pm 0.11}$
& $43.62$ & $14.34_{\pm 0.81}$ & $+1056.45\%$
& $\mathbf{-3.61}$ & $59.84_{\pm 0.86}$ & $+4725.81\%$ \\
 & Qwen3-1.7B
& $85.01$ & $4.13_{\pm 0.11}$
& $52.24$ & $15.15_{\pm 0.33}$ & $+266.83\%$
& $\mathbf{-1.74}$ & $73.53_{\pm 0.70}$ & $+1680.39\%$ \\
\midrule
\multirow{3}{*}{WildGuard} & Gemma-2-2B
& $35.19$ & $1.15_{\pm 0.14}$
& $48.17$ & $5.96_{\pm 0.16}$ & $+418.26\%$
& $\mathbf{13.02}$ & $73.12_{\pm 1.87}$ & $+6258.26\%$ \\
 & Qwen2.5-1.5B
& $70.52$ & $2.86_{\pm 0.05}$
& $45.50$ & $18.78_{\pm 0.76}$ & $+556.64\%$
& $\mathbf{-3.91}$ & $72.19_{\pm 0.75}$ & $+2424.13\%$ \\
 & Qwen3-1.7B
& $83.91$ & $4.81_{\pm 0.33}$
& $65.60$ & $19.37_{\pm 0.48}$ & $+302.70\%$
& $\mathbf{-1.66}$ & $82.34_{\pm 0.38}$ & $+1611.85\%$ \\
\bottomrule
\end{tabular}%
}
\end{table*}

\paragraph{Setup.}
\label{app:llm-setup}
We evaluate CLAMP~\citep{clamp2026} on three language models:
Gemma-2-2B~\citep{gemma2_2024}, Qwen2.5-1.5B~\citep{qwen25_2024}, and Qwen3-1.7B~\citep{qwen3_2025}.
The language-model experiments use
three disjoint splits of SG-Bench~\citep{sgbench2024}: $596$ examples for
constructing the defense, $238$ for optimizing the attack, and $358$ held
out for scoring. For GCG on language models, we optimize a 20-token suffix.
We measure the adaptability $U$ of Eq.~\ref{eq:sgr} via harmful response rate (HR) judged by Llama-Guard-3-1B~\citep{llamaguard2023,llama3_2024} and, on the same responses, by WildGuard~\citep{wildguard2024}.
Similarly to $\mathrm{ACC}_O$ and $\mathrm{ACC}_I$ in Sec.~\ref{sec:exp:metrics}, we report $\mathrm{HR}_O$ and $\mathrm{HR}_I$, the harmful rates of the models adapted from
$M_{\text{prim}}$ and $M_{\text{immu}}$, respectively.
App.~\ref{app:method} and App.~\ref{app:instantiation} give the remaining
implementation details.

\paragraph{Perturbing text in language models.}
\label{app:llm-method}
This paragraph instantiates the model-aware perturbation of
Sec.~\ref{sec:method:a3} for language models.
Unlike images, whose pixels can be optimized directly in a continuous space, text consists of discrete tokens. Given a harmful instruction--response pair \((q,a)\sim\mathcal{D}_{\text{harm}}^{\text{attk}}\), we therefore augment the instruction \(q\) with an optimizable suffix \(s\). The objective is to identify a suffix that induces \(M_{\text{immu}}\) to assign a high likelihood to the harmful response \(a\) conditioned on the augmented instruction \(q\oplus s\):
$
s^{*}
=
\arg\min_{s\in\mathcal{V}^{L}}[
-\lambda\log (1-P_{\text{refusal}})
-\log P_{M_{\text{immu}}}
(a\mid q\oplus s)]$,
where \(\mathcal{V}\) is the token vocabulary, \(L\) is the suffix length, and $P_{\text{refusal}}$ denotes the probability that $M_{\text{immu}}$ refuses to answer, which App.~\ref{app:instantiation} scores at the first generated token. We write \(\log P_{M_{\text{immu}}}(a\mid q\oplus s)\) for the average over the tokens of \(a\), which is truncated to a fixed length, so that this term and the refusal term stay on a comparable scale. Because direct optimization over discrete tokens is infeasible, we use gradients with respect to the token representations to guide the search for \(s^{*}\) like GCG~\citep{gcg2023}. We then replace \((q,a)\) with \((q\oplus s^{*},a)\) during harmful fine-tuning.

\paragraph{Results.}
\label{app:llm-results}
We evaluate CLAMP on the three language models and SG-Bench
splits described above.
The three arms differ only in the
suffix appended to the requests in $\mathcal{D}_{\text{harm}}^{\text{attk}}$:
none (plain adaptation), a GCG-optimized suffix, or the best
of $N$ random suffixes. Tab.~\ref{tab:llm-gcg} shows that, under both judges and plain
adaptation, $\mathrm{HR}_I$ stays between $0.56$ and $4.81$. The GCG suffix raises
$\mathrm{HR}_I$ to between $59.84$ and $82.34$.
GCG-constructed fine-tuning data makes harmful fine-tuning substantially more
effective for both the original and the immunized model, and largely erases
CLAMP's relative protection.

\paragraph{Model-aware versus random suffix.}
As in Sec.~\ref{sec:exp:ablation}, the model-aware perturbation is compared
with a matched control at the same budget. Tab.~\ref{tab:llm-gcg} shows that
the GCG suffix exceeds the best of $N$ random suffixes, drawn at GCG's own candidate
budget and scored by its objective, by between $45.50$ and $67.16$ points in
$\mathrm{HR}_I$ on every model and under both judges.

\paragraph{Original-model harmful rate under the attacks.}
Tab.~\ref{tab:attk-hro} reports the utility of $M_{\text{prim}}$ under each
attack, $\mathrm{HR}_O$ for language models.

\begin{table}[t]
\centering
\caption{$\mathrm{HR}_O$ (\%) of $M_{\text{prim}}$ under the GCG suffix and the random-suffix
control, for the two judges of Tab.~\ref{tab:llm-gcg}.}
\label{tab:attk-hro}
\vspace{3pt}
\footnotesize
\setlength{\tabcolsep}{3pt}
\begin{tabular}{lcccc}
\toprule
& \multicolumn{2}{c}{Llama-Guard-3-1B} & \multicolumn{2}{c}{WildGuard} \\
\cmidrule(lr){2-3}\cmidrule(lr){4-5}
Model & GCG & Random & GCG & Random \\
\midrule
Gemma-2-2B   & $75.29_{\pm 0.91}$ & $11.33_{\pm 3.35}$ & $84.02_{\pm 0.89}$ & $13.16_{\pm 3.72}$ \\
Qwen2.5-1.5B & $57.76_{\pm 0.62}$ & $25.20_{\pm 0.66}$ & $69.49_{\pm 0.60}$ & $34.26_{\pm 0.58}$ \\
Qwen3-1.7B   & $72.32_{\pm 0.98}$ & $31.94_{\pm 1.25}$ & $81.01_{\pm 0.37}$ & $55.90_{\pm 1.20}$ \\
\bottomrule
\end{tabular}
\end{table}

\section{Discussion and limitations}
\label{sec:disc}

\paragraph{What the attacks say together.}
\label{sec:disc:together}
Image misalignment, prompt misalignment, and model-agnostic perturbation require
no access to model gradients: they change only the image source, the textual
description, or the pixels of the attacker's data.  PGA and GCG instead use
gradients to search the attacker's input space, respectively over image pixels
and token suffixes.  Adaptation and model initialization misalignment leave the
data unchanged and instead change the attacker's learning rate or LoRA rank, or rescale the
released attention weights before adaptation.  Despite these different constructions, all
of these results point to the same limitation: a defense evaluated only under the
configuration used during immunization can overstate how much protection remains
after the attacker departs from that configuration.

\paragraph{Complementarity across harmful-data misalignments.}
The generation attacks are not merely interchangeable ways of producing the
same input.  Tab.~\ref{tab:gen-main} reports that the Combined attack, which combines the
substitute generator's images, a name-free style description, and PGA, gives the lowest
$\mathrm{SGR}$ in seven of ten defense--artist settings and ties for lowest
in an eighth, Van~Gogh under \IMMA{}, where prompt misalignment also reaches
$6.46$.  The exceptions are Tyler~Edlin under \IMMA{}, where prompt misalignment
alone reaches $8.98$ while the Combined attack reaches $12.91$, and
Kelly~McKernan under CLAMP, where image misalignment alone reaches
$29.88$ while the Combined attack reaches $31.92$.  We therefore
interpret these misalignments as exposing largely complementary, but not guaranteed
additive, dependencies.

\paragraph{Implications for defense evaluation.}
\label{sec:disc:defender}
A robust evaluation should state which image distribution, caption family,
pixel representation, token context, attacker adaptation strategy, and model
initialization are held fixed.  It should then vary these choices independently before combining
them.  The present results show why reporting only the defense-matched setting
is insufficient: substantial protection remains in some configurations, yet
the measured benefit can shrink sharply when a single attacker-controlled
input is changed.

\paragraph{Evaluation controls.}
\label{sec:disc:method}
For both pixel attacks, the model-agnostic control uses the same perturbation support
and magnitude budget as PGA.  Original-model accuracy is reported alongside
immunized-model accuracy for classification, so a lower $\mathrm{SGR}$ can
be separated from indiscriminate damage to both models.  Prompt misalignment
keeps image bytes unchanged, and image misalignment keeps the concept
and training-set size unchanged.  These controls isolate the input change responsible for each single change to
the harmful data.

\paragraph{Limitations and future work.}
\label{sec:disc:limits}
Extending the evaluation to more artists, concepts, and datasets may help us
better understand how immunization behaves across diverse harmful
capabilities. The same protocol can also incorporate further variations of the
harmful data, the adaptation strategy, and the model initialization to cover
an even wider range of attacker configurations. Such an extended protocol may
also serve as a standard benchmark for future immunization methods.

\end{document}